\documentclass[lettersize,journal]{IEEEtran}
\usepackage{amsmath,amsfonts}
\usepackage{algorithmic}
\usepackage{algorithm}
\usepackage{array}
\usepackage[caption=false,font=normalsize,labelfont=sf,textfont=sf]{subfig}
\usepackage{textcomp}
\usepackage{stfloats}
\usepackage{url}
\usepackage{verbatim}
\usepackage{graphicx}
\usepackage{cite}
\usepackage{hyperref}
\usepackage{booktabs} 
\usepackage{multirow}
\usepackage{subcaption}
\usepackage{bm}
\usepackage{makecell}
\usepackage[table]{xcolor}

\begin{document}

\newcommand{\qz}[1]{{\textcolor{red}{\emph{Qin: #1}}}}
\newcommand{\yijie}[1]{{\textcolor{blue}{\emph{Tang: #1}}}}
\newcommand{\yzn}[1]{{\textcolor{purple}{\emph{Yu: #1}}}}
\newcommand{\revised}[1]{{\textcolor{purple}{#1}}}
\newcommand{\para}[1]{\noindent\textbf{#1}}
\newcommand{\ours}{F2M-Reg}
\newcommand{\todo}[1]{{\textcolor{red}{#1}}}

\title{\ours{}: Unsupervised RGB-D Point Cloud Registration with Frame-to-Model Optimization}

% ---------- todo
\author{
Zhinan~Yu$^{\ast}$, Zheng~Qin$^{\ast}$, Yijie~Tang, Yongjun~Wang, Renjiao~Yi, Chenyang~Zhu, Kai~Xu,~\IEEEmembership{Senior Member,~IEEE}
% <-this % stops a space
\thanks{$^{\ast}$Equal contribution.}
\thanks{This work was supported in part by the National Natural Science Foundation of China under Grants 62325211, 62132021, 62372457 and 62402516, and in part by the Major Program of Xiangjiang Laboratory under Grant 23XJ01009.}
\thanks{\IEEEcompsocthanksitem Zhinan Yu is with the National University of Defense Technology, Changsha 410073, China, and also with the Xiangjiang Laboratory, Changsha 410205, China. 

Yijie Tang, Yongjun Wang, Renjiao Yi, Chenyang Zhu and Kai Xu are with the National University of Defense Technology, Changsha 410073, China. 

Zheng Qin is with the Defense Innovation Institute, Academy of Military Sciences, Beijing 100071, China.}
\thanks{Corresponding author: Kai~Xu (kevin.kai.xu@gmail.com).}
}

\markboth{Journal of \LaTeX\ Class Files,~Vol.~14, No.~8, August~2021}%
{Shell \MakeLowercase{\textit{et al.}}: A Sample Article Using IEEEtran.cls for IEEE Journals}

\IEEEpubid{0000--0000/00\$00.00~\copyright~2021 IEEE}
% ---------- todo

\maketitle

%%%%%%%%%%%%%%%%% ABSTRACT
\begin{abstract}

This work studies the problem of unsupervised RGB-D point cloud registration, which aims at training a robust registration model without ground-truth pose supervision.
Existing methods usually leverages unposed RGB-D sequences and adopt a frame-to-frame framework based on differentiable rendering to train the registration model, which enforces the photometric and geometric consistency between the two frames for supervision.
However, this frame-to-frame framework is vulnerable to inconsistent factors between different frames, e.g., lighting changes, geometry occlusion, and reflective materials, which leads to suboptimal convergence of the registration model.
In this paper, we propose a novel \emph{frame-to-model} optimization framework named \emph{\ours{}} for unsupervised RGB-D point cloud registration.
We leverage the neural implicit field as a global model of the scene and optimize the estimated poses of the frames by registering them to the global model, and the registration model is subsequently trained with the optimized poses.
Thanks to the global encoding capability of neural implicit field, our frame-to-model framework is significantly more robust to inconsistent factors between different frames and thus can provide better supervision for the registration model.
Besides, we demonstrate that \ours{} can be further enhanced by a simplistic synthetic warming-up strategy. To this end, we construct a photorealistic synthetic dataset named \emph{Sim-RGBD} to initialize the registration model for the frame-to-model optimization on real-world RGB-D sequences.
Extensive experiments on four challenging benchmarks have shown that our method surpasses the previous state-of-the-art counterparts by a large margin, especially under scenarios with severe lighting changes and low overlap.
Our code and models are available at \url{https://github.com/MrIsland/F2M_Reg}

\end{abstract}


%%%%%%%%%%%%%%%%% INTRODUCTION
% !Tex root = main.tex
\section{Introduction}
\label{sec:intro}

Point cloud registration is a fundamental but critical task in various 3D vision applications, such as 3D reconstruction, autonomous vehicle, and virtual/augmented reality. Given two partially-overlapped point clouds, it aims at estimating the $6$-DoF relative pose aligning them. In recent years, deep learning-based registration methods~\cite{choy2019fully,huang2021predator,qin2022geometric,yu2023rotation} have attracted great research interests in this area. A typical registration model usually first extracts point features and then retrieves correspondences between two point clouds, based on which the relative pose is computed.
Most learning-based methods are trained in a supervised fashion, which highly depends on high-quality pose annotations. However, current pose annotation methods, either automatic or manual, commonly suffer from unstable convergence and long annotation time, which prevents the usage of large-scale unannotated 3D data and limits the generality of these methods to novel scenes.

\begin{figure*}[t]
\centering
\includegraphics[trim=0.1cm 0.1cm 0cm 0.1cm, clip, width=\linewidth]{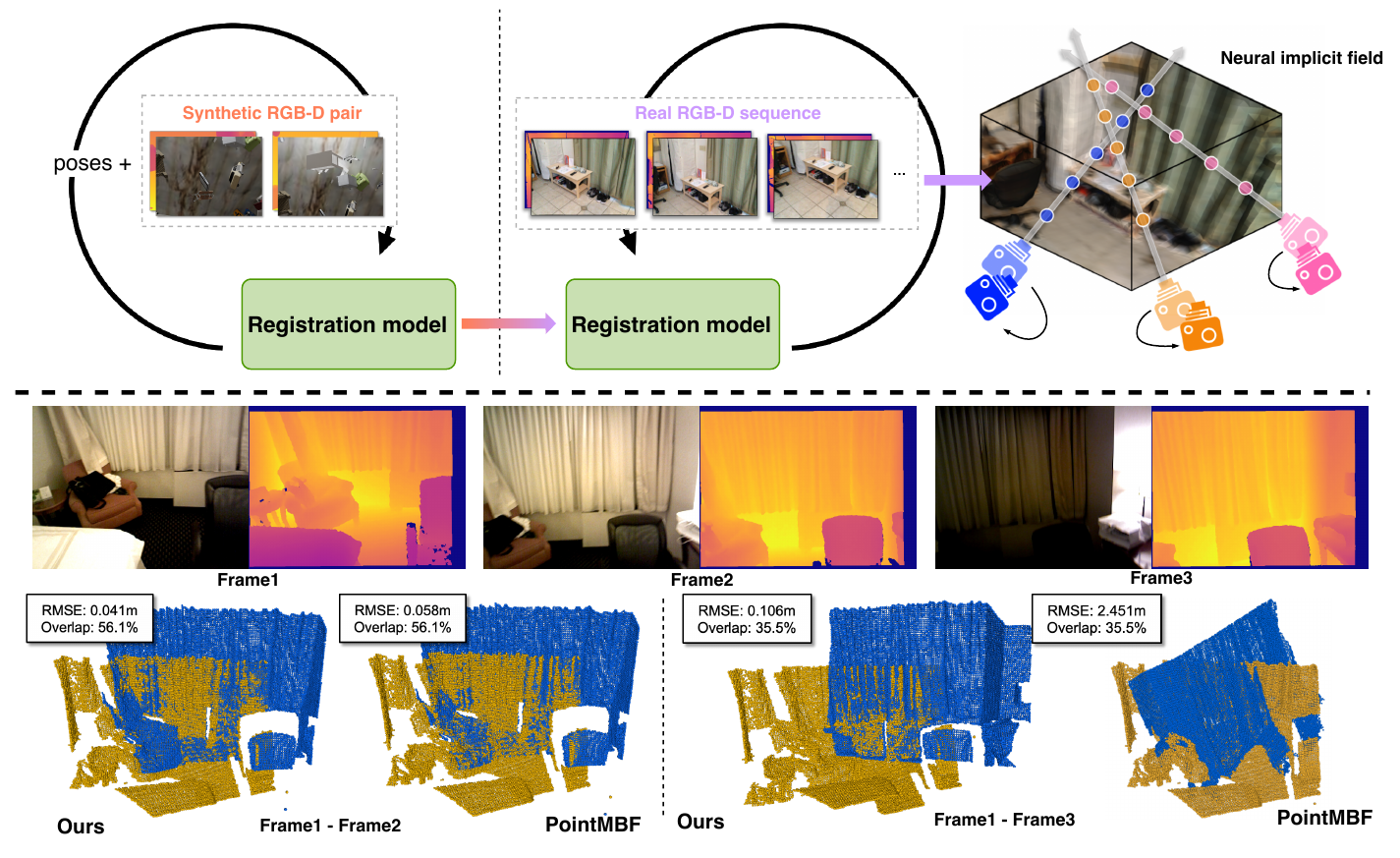}
\caption{
We propose \ours{}, a frame-to-model optimization framework for unsupervised RGB-D registration. The registration is first warmed up with synthetic data, and then fine-tuned on real-world data in the frame-to-model manner (top). Although the frame-to-frame method can successfully register the easy case (bottom-left), it cannot register the case with lighting changes and low overlap (bottom-right). On the contrary, our method effectively register the hard case.
}
\label{fig:setting}
\end{figure*}

For this reason, unsupervised registration methods have been proposed to eliminate the reliance of pose annotations during training. Existing methods~\cite{el2021unsupervisedr, wang2022improving, yuan2023pointmbf} typically leverages unposed RGB-D sequences for training and follow a \emph{frame-to-frame} optimization framework~\cite{el2021unsupervisedr}. The key idea of this framework is to supervise the registration model by enforcing the photometric and geometric consistency between the input RGB-D frames and the rerendered registered frames. 
Given two overlapping RGB-D frames, the registration model first estimates their relative pose. The target frame is then transformed based on the estimated relative pose, and rerendered into the reference frame of the source frame with differentiable rasterization. 
At last, the registration model is trained by enforcing the photometric and geometric consistency between the rerendered target frame and the source frame.
However, the frame-to-frame consistency is vulnerable to inconsistent factors between different frames such as lighting changes, geometry occlusion and reflective materials (see Fig.~\ref{fig:setting}) due to the lack of the global context of the scene.
Furthermore, the differentiable rasterization requires the two frames to have large overlap to generate high-quality rerendering, which limits its application in scenarios with fast motion or severe viewpoint changes.
As a result, the frame-to-frame framework fails to provide effective supervision in these challenging cases, and thus leads to suboptimal convergence of the registration model.

\IEEEpubidadjcol

To address these issues, we propose a novel unsupervised RGB-D point cloud registration framework with \emph{frame-to-model} optimization, named \emph{\textbf{\ours{}}}.
Our key insight is to leverage the powerful global encoding capability of recent neural implicit field~\cite{mildenhall2021nerf,wang2021neus,zhu2022nice,wang2023co} to generate high-quality poses, which are used to supervise the registration model.
Specifically, we maintain a neural implicit field as the global model for each RGB-D sequence. For each frame, we use the registration model to initialize its pose and align it to the global model to further refine the pose. By encoding the radiance and the geometric information with the neural implicit field, the influence of inconsistency across different viewpoints can be effectively alleviated, which allows for more accurate poses and provides better supervision for the registration model.
As shown in Fig.~\ref{fig:setting}, our method still achieves accurate registration in spite of severe lighting changes and low overlap.

It is noteworthy that constructing the neural implicit field requires reasonable initial poses, but the poses from a randomly initialized registration model tend to be of low quality, which thus makes a chicken-and-egg problem.
To this end, we demonstrate that our method can be easily improved with a simplistic synthetic warming-up strategy, where the registration model is first pretrained on synthetic data and then retrained on real-world data with frame-to-model optimization, as shown in Fig.~\ref{fig:setting}.
We create a synthetic dataset named \emph{Sim-RGBD} with photo-realistic rendering of CAD models, which contains more than $100$k rendered images of $90$ scenes.
By warming up the registration model on Sim-RGBD, the quality of the initial poses are significantly improved, which facilitates the further construction of the neural implicit field as well as the frame-to-model optimization.

We conduct extensive experiments on $5$ challenging benchmarks, ScanNet~\cite{dai2017scannet}, 3DMatch~\cite{zeng20173dmatch}, 7-Scenes~\cite{shotton2013scene}, ScanNet++~\cite{Yeshwanth_2023_ICCV} and our Sim-RGBD to evaluate the effectiveness of our method. Our method significantly outperforms previous state-of-the-art methods, especially in more challenging scenarios with lower overlap or severe lighting changes.
For example, without synthetic warming-up, our method surpasses the previous state-of-the-art method~\cite{yuan2023pointmbf} by $6{\sim}14$ percentage points on registration accuracies under the strictest thresholds on ScanNet.
And our method with synthetic warming-up further enlarges the improvements to $10{\sim}17$.
We also conduct comprehensive ablation studies to investigate the design choices of our method.

In summary, our contributions are as follows:
\begin{itemize}
    \item We propose a frame-to-model optimization framework for unsupervised RGB-D point cloud registration guided by a neural implicit field.
    \item We introduce a synthetic warming-up strategy to further improve the performance which provides high-quality initial poses for the frame-to-model optimization.
    \item Our method achieves new state-of-the-art results on several popular RGB-D point cloud registration benchmarks.
\end{itemize}

\begin{comment}
\end{comment}

%%%%%%%%%%%%%%%%% RELATED WORK
% !Tex root = main.tex

\section{Related Work}
\label{sec: related_work}
\subsection{Point Cloud Registration}
\label{subsec: related_pcr}
Point cloud registration is a problem of estimating the transformation matrix between two frames of scanned point clouds. The key lies in how to detect features with specificity from the two-frame point cloud.
Since deep learning has been found good at feature representation, how to learn robust and invariant visual features through deep learning networks has become a focus of research.~\cite{wang2020pie,qi2017pointnet,qi2017pointnet++,duan2022disarm,pan2021variational,qin2022geometric} Many Feature learning methods~\cite{wang2022improving, deng2018ppfnet, gojcic2019perfect, liu2022efficient, ao2021spinnet, bai2020d3feat, choy2020deep, choy2019fully, qin2022geometric, yew2022regtr, yu2023rotation} were proposed. They get the point cloud features by neural network and use a robust estimator e.g. RANSAC to estimate the rigid transformation. Different from focusing on feature learning, there are some end-to-end learning-based registration methods 
~\cite{yang2019extreme, wang2019non, elbaz20173d, lu2019deepvcp, huang2020feature} that treat the registration as a regression problem. They encoded the transformations into the implicit space as a parameter in the network optimization process. 
Recently, 2D foundation models~\cite{rombach2022high, yang2024depth, kirillov2023segment, bhat2023zoedepth} have shown great potential in 3D point cloud registration task. For example, FreeReg~\cite{wang2023freereg} leverages the image features from stable diffusion~\cite{rombach2022high} and estimates the depth information with Zoe-Depth~\cite{bhat2023zoedepth} to solve the image-to-point cloud registration problem.

\subsection{Unsupervised Point Cloud Registration}

The aforementioned methods rely on ground-truth poses to supervised the training. The ground-truth pose is often obtained by reconstruction of the SfM, which suffers from high computational overhead and instability. Recently, unsupervised RGB-D registration methods have been proposed to bypass the need of pose annotations. To our knowledge, UR\&R~\cite{el2021unsupervisedr} is the first unsupervised registration framework by introducing a differentiable render-based loss to optimize the feature extractor. BYOC~\cite{el2021bootstrap} stands for the fact that randomly initialized CNNs also provide relatively good correspondences, proposed a teacher-student framework to train their feature extractor. LLT~\cite{wang2022improving} fused the geometric and visual information in a more trivial way by introducing a multi-scale local linear transformation to fuse RGB and depth modalities. PointMBF~\cite{yuan2023pointmbf} has designed a network based on unidirectional fusion to better extract and fuse features from geometric and visual sources and has achieved state-of-the-art performance. However, these methods have difficulty in handling multi-view inconsistency caused by factors such as lighting changes, highlight or occlusion.

\subsection{Pose Optimization in Neural SLAM}
\label{subsec: nerf_b_p_o}
Existing Neural SLAM methods~\cite{sucar2021imap, zhu2022nice, tang2023mips, wang2023co, zhang2023go, yang2022vox, johari2023eslam} incorporate neural implicit representations into RGB-D SLAM systems, allowing tracking and mapping from scratch. The groundbreaking work, iMAP~\cite{sucar2021imap}, encode both the color and geometry of the scene into a MLP. This MLP can be jointly optimized with a batch of poses through rendering loss. In the subsequent works, NICE-SLAM~\cite{zhu2022nice} and Vox-Fusion~\cite{yang2022vox} introduce a hybrid representation that combines learnable grid-based features with a neural decoder, enabling the utilization of local scene color and geometry to guide pose optimization. More recently, Mipsfusion~\cite{tang2023mips} proposed a robust and scalable RGB-D reconstruction system with a multi-implicit-submap neural representation. Co-SLAM~\cite{wang2023co} proposed a joint coordinate and sparse-parametric encoding and a more global bundle adjustment approach. Inspired by the aforementioned works, we introduce our framework for estimating the initial camera pose using a feature extractor and subsequently refining the pose through implicit 3D reconstruction.

%%%%%%%%%%%%%%%%% METHODS

\section{Method}

\subsection{Overview}

Given two RGB-D frames $\mathcal{X} = (\mathbf{I}^{\mathcal{X}}, \mathbf{X})$ and $\mathcal{Y} = (\mathbf{I}^{\mathcal{Y}}, \mathbf{Y})$, where $\mathbf{I}^{\mathcal{X}}, \mathbf{I}^{\mathcal{Y}} \in \mathbb{R}^{H \times W \times 3}$ are the RGB images and $\mathbf{X} \in \mathbb{R}^{N^{\mathcal{X}} \times 3}, \mathbf{Y} \in \mathbb{R}^{N^{\mathcal{Y}} \times 3}$ are the point clouds backprojected from the corresponding depth images, our goal is to recover the $6$-DoF relative pose $\mathbf{T} \in \mathcal{SE}(3)$ between them, which consists of a 3D rotation $\mathbf{R} \in \mathcal{SO}(3)$ and a 3D translation $\mathbf{t} \in \mathbb{R}^{3}$.
% For simplicity, we denote the point clouds backprojected from the depth images from the two frames as $\mathbf{X} \in \mathbb{R}^{N \times 3}$ and $\mathbf{Y} \in \mathbb{R}^{M \times 3}$.
To solve this problem, a learning-based registration model $\mathcal{F}$ first extracts point features and retrieves point correspondences $\mathcal{C} = \{ (\mathbf{p}_i, \mathbf{q}_i) \mid \mathbf{p}_i \in \mathbf{X}, \mathbf{q}_i \in \mathbf{Y} \}$.
The relative pose is then estimated based on the correspondences. Obviously, the discriminativeness of the extracted features accounts for the quality of the resultant relative pose. However, the training of $\mathcal{F}$ heavily relies on the ground-truth pose $\mathbf{T}^{*} = \{ \mathbf{R}^{*}, \mathbf{t}^{*} \}$, which suffers from great annotation difficulty and unstable convergence.

\begin{figure*}[t]
\centering
\includegraphics[trim=0cm 0.25cm 0.2cm 0.3cm, clip, width=\linewidth]{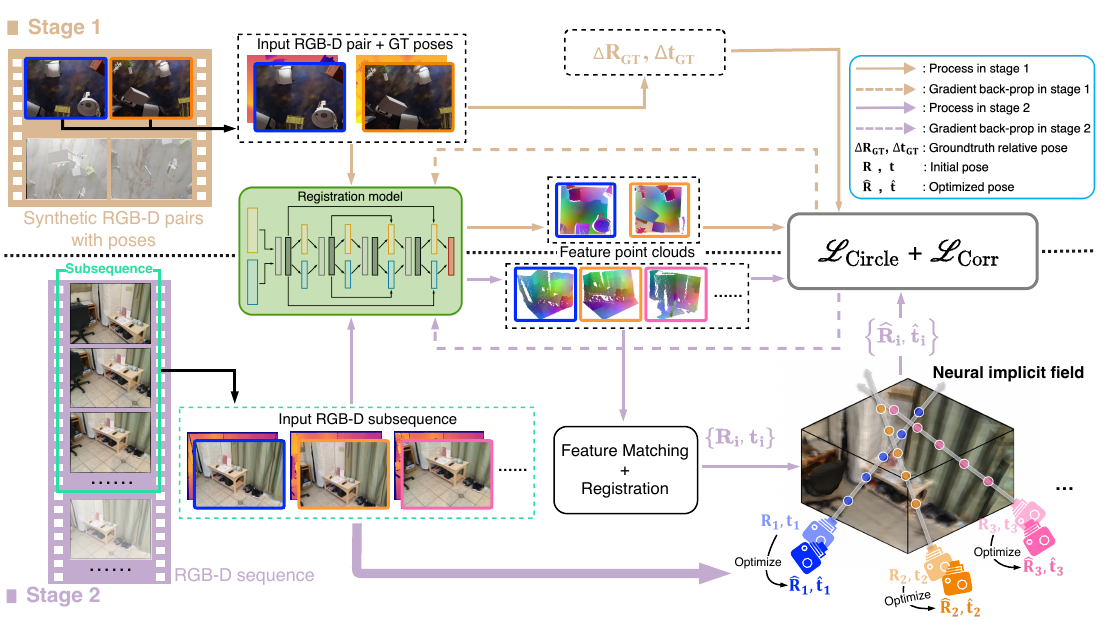}
\caption{Overall pipeline of \ours{}. Our framework can be divided into two stages. The first synthetic warming-up stage leverages synthetic RGB-D pairs as well as their ground-truth poses to train the registration model in a supervised manner. In the second frame-to-model optimization stage, we take an RGB-D sequence as input and use the registration model to estimate the relative pose for every two consecutive frames. Based on the estimated poses, we jointly optimize a neural implicit field of the whole scene and the estimated poses. At last, the optimized poses are used to fine-tune the registration model on real-world data.}
\label{fig:pipeline}
\end{figure*}

In this work, we propose an unsupervised RGB-D point cloud registration method named \emph{\ours{}}. Our method leverages unposed RGB-D sequences to train the registration model $\mathcal{F}$. Fig.~\ref{fig:pipeline} illustrates the overall pipeline of our method. We first describe our registration model in Sec.~\ref{sec:method:basemodel}. To achieve effective supervision, we generates high-quality relative pose in a frame-to-model manner (Sec.~\ref{sec:method:nerf}). To inspire the parameters of $\mathcal{F}$, we build a scene-level synthetic dataset and warm up $\mathcal{F}$ on this dataset so that reasonable initial features can be learned (Sec.~\ref{sec:method:bootstrap}).

\subsection{Registration Model}
\label{sec:method:basemodel}

Our registration model $\mathcal{F}$ adopts a two-branch feature encoder, \emph{i.e.}, the visual branch and the geometric branch, which fuses the information from both the visual (2D) and the geometric (3D) spaces for better feature distinctiveness, similar to PointMBF~\cite{yuan2023pointmbf}. The visual branch is composed of a modified ResNet-$18$~\cite{he2016deep}, following a U-shape architecture. The geometric branch has a KPFCN~\cite{bai2020d3feat, thomas2019kpconv} symmetric with the visual branch. Both branches adopt a three-stage architecture, and a PointNet-based fusion module fuses the features from the two modalities after each stage. The point features of the two frames are, respectively, denoted as $\mathbf{F}^{\mathcal{X}} \in \mathbb{R}^{N^{\mathcal{X} \times C}}$ and $\mathbf{F}^{\mathcal{Y}} \in \mathbb{R}^{N^{\mathcal{Y} \times C}}$, which are $\ell_2$-normalized onto a unit hypersphere.
We then extract the correspondences by mutual matching based on the point features, and select the top-$N_C$ correspondences with the smallest distances in the feature space. At last, the correspondences are fed into RANSAC to estimate the final relative pose.

\subsection{Unsupervised Registration with Frame-to-Model Optimization}
\label{sec:method:nerf}

An unsupervised registration pipeline usually first gives a rough estimation of the pose, and then supervise the registration model with the estimated pose.
Obviously, the quality of the estimated pose significantly affects the accuracy of the trained registration model.
Existing methods~\cite{el2021unsupervisedr,wang2022improving,yuan2023pointmbf} use differentiable rasterization and optimize the frame pose according to the photometric and the geometric consistency between two nearby frames from an RGB-D sequence. Nevertheless, the \emph{frame-to-frame} consistency can be easily affected by occlusion or lighting changes under different viewpoints, which leads to suboptimal poses and thus harms the training of the registration model. This has inspired us that a more comprehensive modeling of the whole scene is required to effectively optimize the frame poses, \emph{i.e.}, \emph{frame-to-model} optimization.
Recently, the neural implicit field~\cite{mildenhall2021nerf, wang2021neus} has shown a strong ability to model appearance and geometric structures in a scene, and jointly optimize 3D maps and poses~\cite{zhu2022nice,wang2023co}.
Based on this insight, we propose to train the registration model scene by scene and optimize a neural implicit field as a global model for each scene for better pose refinement. 
This allows to optimize the poses in a frame-to-model fashion instead of the traditional frame-to-frame one, which can better handle the occlusion and lighting changes.

\para{Training pipeline.}
As shown in Fig.~\ref{fig:pipeline}, to avoid the error accumulation and the huge time overhead caused by joint map-pose optimization in long sequences, we opt to process small subsquences instead of the whole sequence.
Specifically, we split the RGB-D sequence of a scene into subsequences of $200$ frames, and we optimize a neural implicit field $\mathcal{M}$ for each subsequence. Within each subsequence, we further sample keyframes every $20$ frames for training and all other frames are omitted. The reference frame of the first keyframe is treated as the anchor reference frame of the subsquence. For each keyframe, we first register it with the previous keyframe with $\mathcal{F}$ to obtain its initial pose, and then insert it into $\mathcal{M}$ to jointly optimize its pose and the map. 
At last, we use the optimized pose of each keyframe to supervise the registration model.

\para{Initial pose generation.}
We use a method similar to~\cite{el2021unsupervisedr} to generate the initial pose.
Given the point features $\mathbf{F}^{\mathcal{X}}$ and $\mathbf{F}^{\mathcal{Y}}$, for each point $\mathbf{x}_i \in \mathbf{X}$, we then find its nearest point $\mathbf{y}_{n_i} \in \mathbf{Y}$ in the feature space as a correspondence. The weight for each correspondence is computed as:
\begin{equation}
w_{i} = 1 - \frac{\lVert \mathbf{f}^{\mathcal{X}}_{i} - \mathbf{f}^{\mathcal{Y}}_{n_i} \rVert}{2}.
\end{equation}
At last, we select the top $k$ correspondences with the largest weights. The same computation goes for $\mathbf{Y}$. As a result, we obtain $2k$ correspondences, denoted as $\mathcal{C}$. To compute the initial pose, we randomly sample $t$ correspondence subsets, each containing $r$ of the correspondences. For each subset, we use weighted SVD~\cite{besl1992method} to compute a pose hypothesis and select the best pose which minimizes:
\begin{equation}
E = \sum_{(\mathbf{p}_i, \mathbf{q}_i) \in \mathcal{C}} w_{i} \lVert \mathbf{R} \mathbf{p}_i + \mathbf{t} - \mathbf{q}_i \rVert.
\label{eq:correspondence-error}
\end{equation}

\begin{figure*}[t]
\centering
\includegraphics[trim=0.cm 0.1cm 0.cm 0.1cm, clip, width=\linewidth]{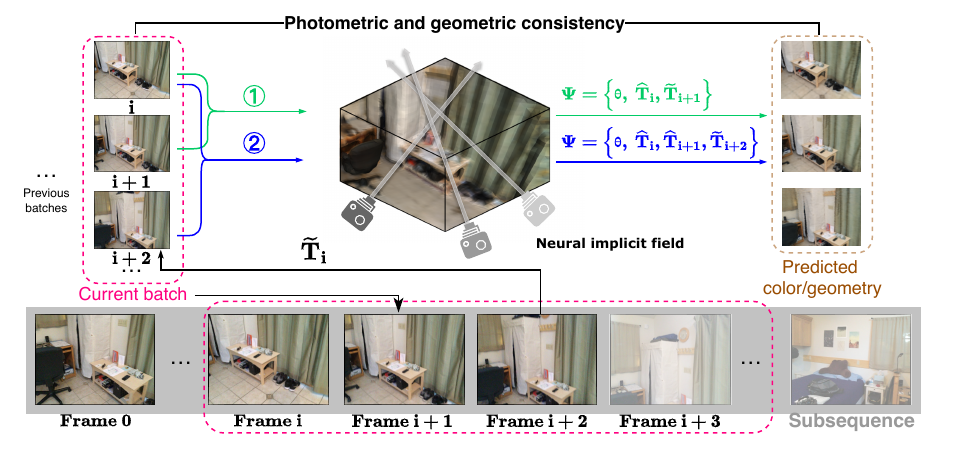}
\caption{Mapping stage. 
The first frame in the current batch, Frame $i$, can either be the first of a new subsequence or the last from the previous batch, with its pose known. Once a new frame is tracked, it is added to the batch with its tracked pose $\mathbf{\tilde{T}}_{i}$. In Step 1, when the $(i+1)^\text{th}$ frame is added, its tracked pose $\mathbf{\tilde{T}}_{i+1}$ is optimized along with the mapped pose $\mathbf{\hat{T}}_{i}$ and the implicit scene representation $\theta$. In Step 2, after adding the $(i+2)^\text{th}$ frame, the optimization parameters become $\mathbf{\Psi} = \{\theta, \mathbf{\hat{T}}_{i}, \mathbf{\hat{T}}_{i+1}, \mathbf{\tilde{T}}_{i+2}\}$, with further frames tracked and optimized similarly.
}
\label{fig:BA}
\end{figure*}

\para{Pose optimization.}
We adopt a neural implicit field similar with Co-SLAM~\cite{wang2023co} due to its advances in the speed and the quality of reconstruction. Our neural field maps the world coordinates $\mathbf{x} = (x, y, z)$ and the viewing direction $\mathbf{d} = (\theta, \phi)$ into the color $\mathbf{c}$ and the TSDF value $s$. Following the SLAM pipeline, for each keyframe, our method is split into the \emph{tracking} stage and \emph{mapping} stage.

In the tracking stage, we optimize the pose of the keyframe with the neural implicit field $\mathcal{M}$. For clarity, the optimized pose in this stage is named the \emph{tracked pose}, denoted as $\tilde{\mathbf{T}}_i$. 
For the $i$-th keyframe, we first calculate its untracked pose $\mathbf{T}_i = \Delta\mathbf{T}_{i-1,i} \cdot \hat{\mathbf{T}}_{i-1}$, where $\hat{\mathbf{T}}_{i-1}$ is the mapped pose of the previous keyframe as described later and $\Delta\mathbf{T}_{i-1,i}$ is their initial relative pose from the registration model. $\mathbf{T}_i$ is then optimized to $\tilde{\mathbf{T}}_i$ by supervising the photometric and the geometric consistency between the input RGB-D frame and the rerendered frame by $\mathcal{M}$. The neural implicit field $\mathcal{M}$ is fixed in this stage. 
As the filed implicitly models the whole scene, this frame-to-model paradigm could alleviate the influence of heavy occlusion or lighting changes from different viewpoints, and thus achieves more effective optimization of the keyframe pose.

After one keyframe is tracked, we jointly optimize the $\mathcal{M}$ and the poses of the keyframes in the mapping stage. The pose refined in this stage is named the \emph{mapped pose}, denoted as $\hat{\mathbf{T}}_i$.
The mapping stage adopts a batch-wise optimization strategy.
When a keyframe is tracked, it is added into the current batch. Then all keyframes in this batch are used to optimize the $\mathcal{M}$ to improve the implicit scene model, with their poses being optimized simultaneously. This joint optimization further improves the quality of the keyframe poses.
After we have collected a batch of keyframes with a maximal size $B$, we train the registration model with the mapped poses of the keyframes in the current batch. The batch is then emptied except the last keyframe, which is used to provides the anchor pose for the coming keyframes in the next batch.

\para{Neural implicit field optimization.}
Building on Co-SLAM~\cite{wang2023co}, we optimize poses and the neural implicit field by minimizing five distinct losses across both the tracking and mapping stages:

(1) Two rendering losses $\mathcal{L}_{\text{rgb}}$ and $\mathcal{L}_{\text{depth}}$ for minimizing errors between ground truth RGB/depth image $\hat{C}_p$/$\hat{D}_p$ and rendered RGB/depth image $C_p$/$D_p$:
\begin{equation}
\begin{aligned}
& \mathcal{L}_{\text{rgb}}=\frac{1}{|P|} \sum_{p\in P} (C_p-\hat{C}_p)^2, \\ &  \mathcal{L}_{\text{depth}}=\frac{1}{|P|} \sum_{p\in P} (D_p-\hat{D}_p)^2,    
\end{aligned}
\label{eq:rgb_depth_loss}
\end{equation}
where $P$ represents sampled image pixels.

(2) An SDF loss $\mathcal{L}_{\text{sdf}}$ to enhance the consistency of the SDF field:
\begin{equation}
\mathcal{L}_{\text{sdf}}=\frac{1}{|P|} \sum_{p\in P} \frac{1}{|S_p^{\text{tr}}|} \sum_{s\in S_p^{\text{tr}}} (D_s-\hat{D}_s)^2,
\end{equation}
$S_p^{tr}$ represents whose signed distance function (SDF) is not truncated along the viewing ray of pixel $p$, and $D_s$/$\hat{D}_s$ denote their predicted/ground-truth SDF values.

(3) For those sampled points distant from the observed surface, a free-space loss $\mathcal{L}_{\text{fs}}$ is applied to enforce their predicted SDF to be truncation distance $d_{\text{tr}}$: 
\begin{equation}
\mathcal{L}_{\text{fs}}=\frac{1}{|P|} \sum_{p\in P} \frac{1}{|S_p^{\text{fs}}|} \sum_{s\in S_p^{\text{fs}}} (D_s-d_{\text{tr}})^2.
\end{equation}

(4) An additional regularization on the interpolated features $\mathcal{V}_{\alpha}(x)$ in order to decrease the noisy in reconstruction.
\begin{equation}
    \mathcal{L}_{\text{smooth}} = \frac{1}{\mathcal{V}}\sum_{x\in\mathcal{|V|}} (\Delta_{x}^{2} +\Delta_{y}^{2} + \Delta_{z}^{2})
\end{equation}
where $\mathcal{V}$ denotes the grid and $\Delta_{\text{xyz}} = \mathcal{V}_{\alpha}(x + \epsilon_{\text{xyz}}) - \mathcal{V}_{\alpha}(x)$

The overall loss is computed as:
\begin{equation}
    \mathcal{L} = \lambda_{\text{rgb}} \mathcal{L}_{\text{rgb}} + \lambda_{\text{depth}} \mathcal{L}_{\text{depth}} + \lambda_{\text{sdf}} \mathcal{L}_{\text{sdf}} + \lambda_{\text{fs}} \mathcal{L}_{\text{fs}} + \lambda_{\text{smooth}} \mathcal{L}_{\text{smooth}},
\end{equation}
where the weights of each loss are $\lambda_{\text{rgb}} = 5.0$, $\lambda_{\text{depth}} = 0.1$, $\lambda_{\text{sdf}} = 1000$, $\lambda_{\text{fs}} = 10$, $\lambda_{\text{smooth}} = 0.001$.

\para{Training the registration model. }
After obtaining the optimized poses of a batch, we compute the relative poses between consecutive keyframes to train the registration model. Specifically, the relative pose is computed as $\Delta \mathbf{T}_{i-1, i} = \hat{\mathbf{T}}_i \hat{\mathbf{T}}_{i-1}^{-1}$. we then extract correspondences between two keyframes based on $\Delta \mathbf{T}_{i-1, i}$. The correspondences between two point clouds are identified using a specified threshold $\tau$, which can be formulated as $\mathcal{C}^{*} = \{(\mathbf{p}_{i-1}, \mathbf{p}_{i}) \mid ||\mathbf{T}_{i-1, i}\mathbf{p}_{i-1} - \mathbf{p}_{i}|| < \tau\}$. And we apply the circle loss~\cite{sun2020circle,huang2021predator} and the correspondence loss (Eq.~\ref{eq:correspondence-error}) on the extracted correspondences to train the registration model.

The correspondence loss $\mathcal{L}_{\text{corr}}$ (1) is formalized in Eq.~\ref{eq:correspondence_loss}.  We choose the top 256 pairs of correspondences and use the relative optimized pose $\Delta \mathbf{\hat{T}} = [ \, \mathbf{\hat{R}} \, | \,  \mathbf{\hat{t}} \, ]$ to calculate the loss. The weights $w_{i}$ range from 0 to 1, and are derived from the cosine similarity values of the two point features.
\begin{equation}
\mathcal{L}_{\text{corr}} = \sum_{(\mathbf{p}_i, \mathbf{q}_i) \in \mathcal{C}} w_i \lVert \mathbf{\hat{R}} \mathbf{p}_i + \mathbf{\hat{t}} - \mathbf{q}_i \rVert
\label{eq:correspondence_loss}
\end{equation}

The circle loss $\mathcal{L}_{\text{circle}}$ is formalized in Eq.~\ref{eq:circle_loss}. Considering the correspondence $\mathcal{C} = \{ (\mathbf{p}_i, \mathbf{q}_i) \mid \mathbf{p}_i \in \mathbf{X}, \mathbf{q}_i \in \mathbf{Y} \}$ and the optimized pose $\hat{T}$. We compute, for each point in $\mathbf{X}$ the distance to all points in $\mathbf{Y}$. Pairs of points with a distance less than $r_{p}$ are treated as positive samples $\epsilon_{\text{pos}}$, while those greater than $r_{s}$ are treated as negative samples $\epsilon_{neg}$. The circle loss from $\mathbf{X}$ is formalized in Eq.~\ref{eq:circle_loss}.
\begin{equation}
\begin{aligned}
\mathcal{L}_{\text{circle}}^{\mathbf{X}} = \frac{1}{n} \sum_{i=1}^{n} \log [ 1 + & \sum_{j \in \epsilon_{\text{pos}}} e^{\beta_{\text{pos}}^{j} (d_{i}^{j} - \Delta_{\text{pos}}) } \\ & \mathbf{\cdot} \sum_{k \in \epsilon_{\text{neg}} } e^{\beta_{\text{neg}}^{k} (\Delta_{\text{neg}} - d_{i}^{k}) } ]
\label{eq:circle_loss}
\end{aligned}
\end{equation}
where $n$ is the number of the points in $\mathbf{X}$, $d_{i}^{j} = \lVert f_{p_{i}} - f_{q_{i}} \lVert$ denotes the L2 distance of the corresponding point features and $\Delta_{\text{pos}}$, $\Delta_{\text{neg}}$ are positive and negative margins. The weights $\beta_{\text{pos}}^{j}  = \gamma (d_{i}^{j} - \Delta_{\text{pos}})$ and $\beta_{\text{neg}}^{k} = \gamma (\Delta_{\text{neg}} - d_{i}^{k}) $ are computed for each correspondence. The margin hyper-parameters are set to $\Delta_{\text{pos}} = 0.1$ and $\Delta_{neg} = 1.4$. For the circle loss $\mathcal{L}_{\text{circle}}^{\mathbf{Y}}$ goes the same. The final circle loss is computed as $\mathcal{L}_{\text{circle}} = (\mathcal{L}_{\text{circle}}^{\mathbf{X}} + \mathcal{L}_{\text{circle}}^{\mathbf{Y}}) / 2$. The final loss is the sum of the above two loss terms: $\mathcal{L} = \mathcal{L}_{\text{corr}} + \mathcal{L}_{\text{circle}}$.

\subsection{Synthetic Warming-up}
\label{sec:method:bootstrap}

Jointly optimizing the neural implicit field $\mathcal{M}$ and the poses of keyframes requires relatively accurate initial poses. However, a randomly initialized registration model tends to generate enormous outlier correspondences. This causes the initial poses to be erroneous, and thus leads to suboptimal convergence. To address this issue, we devise a synthetic warming-up strategy which leverages synthetic data to initialize the registration model. The advantages are two-fold. On one hand, with the synthetic data, we can warm up the registration model under the supervision of the ground-truth poses so that it can provide reasonable initial poses. On the other hand, it is extremely easy to scale up the synthetic data and obtain a large-scale dataset for model training.

\vspace{10pt}
\para{Sim-RGBD dataset.}
To warm up the model training, we first construct a synthetic dataset named \emph{Sim-RGBD} using photo-realistic simulation with BlenderProc~\cite{denninger2023blenderproc2}. Sim-RGBD consists of $90$ scenes, which are split into $60$ training scenes and $30$ validation scenes.
% We use the object models from ShapeNet dataset~\citep{chang2015shapenet} to compose the scenes. 
Specifically, for each scene, we create two boxes centered at $(0, 0, 0)$ in the sizes of respectively $10\text{m} \times 10\text{m} \times 5\text{m}$ and $6\text{m} \times 6\text{m} \times 3\text{m}$, and uniformly select $400$ positions in the space between them. We then place a random object model from ShapeNet dataset~\cite{chang2015shapenet} at each position, which is randomly rotated, translated, and scaled. For simplicity, we do not detect collisions when placing the objects.

\begin{figure}[]
\centering
\includegraphics[width=\linewidth]{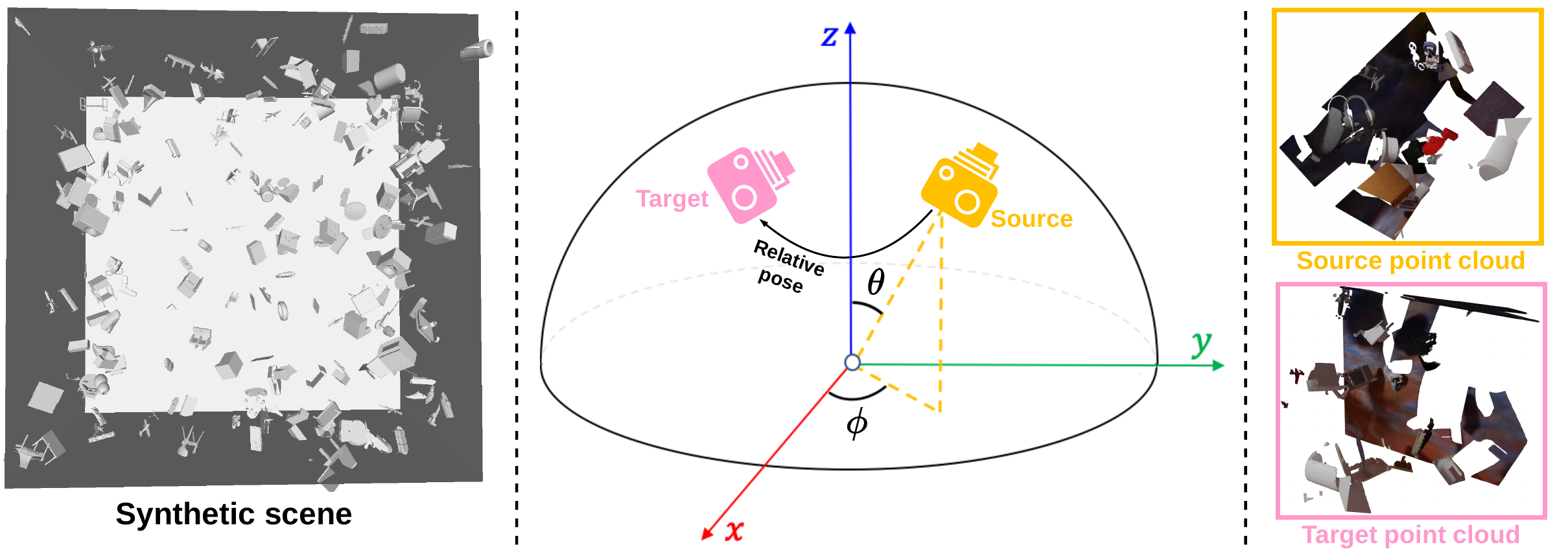}
\caption{\textbf{Demonstration of Sim-RGBD dataset.} The entire scene is depicted in the left figure. Camera sampling is illustrated in the middle figure. Initially, we sample the position of the first camera based on a specified pitch angle $\theta$ and yaw angle $\phi$, with (0, 0, 0) as the viewpoint, forming the camera's view direction. The position of the second camera is derived from the transformation of the first camera position, which is obtained from a Gaussian distribution. The right figure showcases the point cloud with color extracted from the scene.}
\label{fig:sim}
\end{figure}

After constructing the synthetic scenes, we render $400$ pairs of RGB-D frames from each scene. 
To mimic the realistic distribution of the camera poses, we first sample the pose of the source frame, and then sample the relative pose between the source and the target frames. For the source pose, the camera direction is determined by a random pitch angle between $[15^{\circ}, 75^{\circ}]$ and a random yaw angle between $[0^{\circ}, 360^{\circ}]$. And the camera position is determined by a random distance between $[0.7\text{m}, 1.5\text{m}]$ from $(0, 0, 0)$ along this direction. For the relative pose, we first randomly sample a rotation axis, and then sample the rotation angle from $\mathcal{N}(20^{\circ}, 15^{\circ})$ and the translation from $\mathcal{N}(0.4\text{m}, 0.2\text{m})$.
As the two rendered frames could have little overlap, we only preserve the pairs with the overlap ratio above $0.3$. As shown in Sec.~\ref{subsec: abl_study}, the synthetic scenes simulated with this simplistic strategy effectively warm up the model. Fig.~\ref{fig:sim} visualizes the process of sampling two camera poses.

\vspace{10pt}
\para{Training settings.}
Similar to Sec.~\ref{sec:method:nerf}, we use the ground-truth poses to retrieve correspondences and apply the circle loss to train the registration model. Note that it is important to ensure the warming-up stage and the frame-to-model optimization stage to use the same feature formulation, e.g., $\ell_2$ normalized. Otherwise, the two models could be in different feature spaces, thus harming the final performance.

%%%%%%%%%%%%%%%%% RESULTS
% !Tex root = main.tex

\section{Experiments}
\label{sec: experiments}
We evaluate the effectiveness of our method on $5$ RGB-D datasets: 3DMatch~\cite{zeng20173dmatch}, ScanNet~\cite{dai2017scannet}, ScanNet++~\cite{Yeshwanth_2023_ICCV}, 7 Scenes~\cite{shotton2013scene} and our Sim-RGBD. We first describe the implementation details in Sec.~\ref{subsec: experimental_setup}. Next, we compare our method with the baselines in Sec.~\ref{subsec: ex_res} and Sec.~\ref{subsec: ex_res_tum}. At last, we perform comprehensive ablation studies in Sec.~\ref{subsec: abl_study}.

\subsection{Implementation Details}
\label{subsec: experimental_setup}

\begin{table}[]
    \centering
     \resizebox{\linewidth}{!}{
    \begin{tabular}{l|l}
    \hline
        Momentum & 0.9 \\
        Optimizer & Adam \\
        Image size & 128 * 128 \\
        Feature dimension & 32 \\
        $K_{v2g}$,$K_{g2v}$ for training & $K_{v2g} = 16$,$K_{g2v} = 1$ \\
        $K_{v2g}$,$K_{g2v}$ for testing & $K_{v2g} = 32$,$K_{g2v} = 1$ \\
        \hline
         \rowcolor[HTML]{EFEFEF} Batch size & 4 \\
         \rowcolor[HTML]{EFEFEF} Normalization in ResNet & GroupNorm \\
         \rowcolor[HTML]{EFEFEF} Normalization in fusion & GroupNorm \\
         \rowcolor[HTML]{EFEFEF} Group of channels & 32\\
         \rowcolor[HTML]{EFEFEF} Number of correspondence $k$ & 512 \\
    \hline
    \end{tabular}}
    \caption{Implementation details of our \ours{}}
    \label{tab:implementation}
\end{table}

% \para{Implementation.} 
\para{Network architecture.}
The overall network framework we follow is based on PointMBF~\cite{yuan2023pointmbf}. As described in~\ref{sec:method:basemodel}, the network employs a two-branch feature encoder.
The visual branch adopts a $3$-stage U-Net architecture based on ResNet, which accepts images with a resolution of $128 \times 128$ pixels as input. In the encoder, the feature channel starts from 64, and is doubled after each downsampling. 
In the decoder, we use bilinear upsampling to increase the resolution of the feature maps.
The geometric branch adopts a $3$-stage KPFCN~\cite{thomas2019kpconv}. The point clouds are voxelized by $2.5$cm before being fed into the network and are downsampled by $2$ using grid subsampling after each stage. In the encoder, the feature dimension is initially 128 and gets doubled after each downsampling. In the decoder, nearest-neighbor upsampling is used.
Both branches use GroupNorm with $32$ groups and ReLU as the activation.
At last, the features from the two branches are projected to $32$ channels with two respective linear projections to unify the dimension of the final features.

\para{Training and testing.}
We implement and evaluate \ours{} using Pytorch~\cite{imambi2021pytorch} and PyTorch3D~\cite{ravi2020pytorch3d} on an RTX 3090Ti GPU. The registration models are trained using the Adam optimizer~\cite{kingma2014adam} for 20 epochs on Sim-RGBD, 2 epochs on 3DMatch and 1 epoch on ScanNet. The batch size is 4 and the weight decay is $10^{-6}$. The learning rate starts from $10^{-4}$ and decays exponentially by $0.1$ every epoch. For the initial pose generation, we select top $k=256$ correspondences for each side, and use $t=10$ and $r=20\%$. 

\para{Neural implicit field.} In each optimization step, we randomly select $N_{t} = 1024$ pixels from each keyframe. For each ray, we uniformly sample $M_{c} = 32$ points between the near and far bounds. Additionally, we sample an extra $M_{f} = 21$ depth-guided points evenly within the range $[d - d_{s}, d + d_{s}]$, where $d$ represents the depth and $d_{s}=0.25$ denotes a small offset.

\para{Metrics.} Following prior work~\cite{yuan2023pointmbf, wang2022improving, el2021unsupervisedr}, we evaluate our \ours{} method using three key metrics: (1) \emph{Relative Rotation Error} (RRE), which measures the geodesic distance between the estimated and ground-truth rotation matrices; (2) \emph{Relative Translation Error} (RTE), which quantifies the Euclidean distance between the estimated and ground-truth translation vectors; (3) \emph{Chamfer Distance} (CD), defined as the sum of squared mutual minimum Euclidean distances between the registered point clouds of the estimated and ground-truth alignments; and (4) \emph{Inlier Ratio} (IR), the fraction of putative correspondences whose residuals fall below a specified threshold (e.g., 0.1m) when evaluated using the ground-truth transformation. RRE, RTE, and CD are reported at three different thresholds, with both mean and median values provided, while IR is reported as a percentage.

\para{Baseline Methods. }
We compare our method with the baselines from three categories, (1) traditional methods: ICP~\cite{besl1992method}, FPFH~\cite{rusu2009fast} and
SIFT~\cite{lowe2004distinctive}, (2) supervised learning-based methods: SuperPoint~\cite{detone2018superpoint}, FCGF~\cite{choy2019fully}, DGR~\cite{choy2020deep}, 3D MV Reg~\cite{gojcic2020learning} and REGTR~\cite{yew2022regtr}, and (3) unsupervised methods: UR\&R~\cite{el2021unsupervisedr}, BYOC~\cite{el2021bootstrap}, LLT~\cite{wang2022improving} and PointMBF~\cite{yuan2023pointmbf}.

\subsection{Evaluations on ScanNet}
\label{subsec: ex_res}

\para{Dataset.}
ScanNet~\cite{dai2017scannet} and 3DMatch~\cite{zeng20173dmatch} are two indoor RGB-D datasets for point cloud registration task. 
ScanNet consists of $1457$ scenes, with $1045$ for training, $312$ for validation and $100$ for testing. And 3DMatch consists of $101$ scenes, with $71$ for training, $19$ for validation and $11$ for testing.
In this section, we train the models separately on two datasets and evaluate them on the testing split of ScanNet.

During training, we divide each sequence into subsequences with $200$ frames, resulting in about $8$K training subsequences on ScanNet and about $0.5$K subsequences on 3DMatch.
For each subsequence, we sample a keyframe every $20$ frames.
Note that we use less training data than previous work~\cite{el2021unsupervisedr,wang2022improving,yuan2023pointmbf} as we do not exhaustively enumerate all frame pairs.

During testing, we follow the strategy in~\cite{el2021unsupervisedr} to sample the testing pairs, but in two settings: (1) \emph{$\mathit{20}$-frame setting} as in~\cite{el2021unsupervisedr,wang2022improving,yuan2023pointmbf}, where the two frames in each pair are $20$ frames apart, and (2) \emph{$\mathit{50}$-frame setting}, where the two frames in each pair are $50$ frames apart.
The $50$-frame setting is more challenging due to more significant viewpoint variations, which is more consistent with real-world applications.

\begin{table*}[t]
    \centering
    \small
     \setlength{\tabcolsep}{3.0pt}
    {\begin{tabular}{l|cc|ccccc|ccccc|ccccc}
    \toprule
       \multicolumn{1}{l|}{\multirow{3}{*}{ }} & \multicolumn{1}{c}{\multirow{3}{*}{Train Set}} & \multicolumn{1}{c|}{\multirow{3}{*}{Sup}} & \multicolumn{5}{c|}{Rotation($^{\circ}$)} & \multicolumn{5}{c|}{Translation(cm)} & \multicolumn{5}{c}{Chamfer(mm)} \\
       \multicolumn{1}{l|}{} & \multicolumn{1}{c}{} & \multicolumn{1}{c|}{} & \multicolumn{3}{c}{Accuracy $\uparrow$}  & \multicolumn{2}{c|}{Error$\downarrow$} & \multicolumn{3}{c}{Accuracy $\uparrow$} & \multicolumn{2}{c|}{Error$\downarrow$} & \multicolumn{3}{c}{Accuracy $\uparrow$} & \multicolumn{2}{c}{Error$\downarrow$} \\
        \multicolumn{1}{l|}{} & \multicolumn{1}{c}{} & \multicolumn{1}{c|}{} & 5 & 10 & 45 & Mean & Med. & 5 & 10 & 25 & Mean & Med. & 1 & 5 & 10 & Mean & Med. \\
    \midrule
        ICP~\cite{besl1992method} & - & & 31.7 & 55.6 & \underline{99.6} & 10.4 & 8.8 & 7.5 & 19.4 & 74.6 & 22.4 & 20.0 & 8.4 & 24.7 & 40.5 & 32.9 & 14.1 \\
        FPFH~\cite{rusu2009fast} & - & & 34.1 & 64.0 & 90.3 & 20.6 & 7.2 & 8.8 & 26.7 & 66.8 & 42.6 & 18.6 & 27.0 & 60.8 & 73.3 & 23.3 & 2.9 \\
        SIFT~\cite{lowe2004distinctive} & - & & 55.2 & 75.7 & 89.2 & 18.6 & 4.3 & 17.7 & 44.5 & 79.8 & 26.5 & 11.2 & 38.1 & 70.6 & 78.3 & 42.6 & 1.7 \\
        SuperPoint~\cite{detone2018superpoint} & - & & 65.5 & 86.9 & 96.6 & 8.9 & 3.6 & 21.2 & 51.7 & 88.0 & 16.1 & 9.7 & 45.7 & 81.1 & 88.2 & 19.2 & 1.2 \\
    \midrule
        FCGF~\cite{choy2019fully} & 3DMatch$^{*}$ & \checkmark & 70.2 & 87.7 & 96.2 & 9.5 & 3.3 & 27.5 & 58.3 & 82.9 & 23.6 & 8.3 & 52.0 & 78.0 & 83.7 & 24.4 & 0.9 \\
        DGR~\cite{choy2020deep} & 3DMatch$^{*}$ & \checkmark & 81.1 & 89.3 & 94.8 & 9.4 & 1.8 & 54.5 & 76.2 & 88.7 & 18.4 & 4.5 & 70.5 & 85.5 & 89.0 & 13.7 & 0.4 \\
        3D MV Reg~\cite{gojcic2020learning} & 3DMatch$^{*}$ & \checkmark & 87.7 & 93.2 & 97.0 & 6.0 & 1.2 & 69.0 & 83.1 & 91.8 & 11.7 & 2.9 & 78.9 & 89.2 & 91.8 & 10.2 & \underline{0.2} \\
        REGTR~\cite{yew2022regtr} & 3DMatch$^{*}$ & \checkmark & 86.0 & 93.9 & 98.6 & 4.4 & 1.6 & 61.4 & 80.3 & 91.4 & 14.4 & 3.8 & 80.9 & 90.9 & 93.6 & 13.5 & \underline{0.2} \\
        UR\&R (Supervised) & 3DMatch & \checkmark & 92.3 & 95.3 & 98.2 & 3.8 & \textbf{0.8} & 77.6 & 89.4 & 95.5 & 7.8 & 2.3 & 86.1 & 94.0 & 95.6 & 6.7 & \textbf{0.1} \\
    \midrule
        UR\&R~\cite{el2021unsupervisedr} & 3DMatch & & 87.6 & 93.1 & 98.3 & 4.3 & 1.0 & 69.2 & 84.0 & 93.8 & 9.5 & 2.8 & 79.7 & 91.3 & 94.0 & 7.2 & \underline{0.2} \\
        UR\&R (RGB-D) & 3DMatch & & 87.6 & 93.7 & 98.8 & 3.8 & 1.1 & 67.5 & 83.8 & 94.6 & 8.5 & 3.0 & 78.6 & 91.7 & 94.6 & 6.5 & \underline{0.2} \\
        BYOC~\cite{el2021bootstrap} & 3DMatch & & 66.5 & 85.2 & 97.8 & 7.4 & 3.3 & 30.7 & 57.6 & 88.9 & 16.0 & 8.2 & 54.1 & 82.8 & 89.5 & 9.5 & 0.9 \\
        LLT~\cite{wang2022improving} & 3DMatch & & 93.4 & 96.5 & 98.8 & 2.5 & \textbf{0.8} & 76.9 & 90.2 & 96.7 & 5.5 & \underline{2.2} & 86.4 & 95.1 & 95.8 & 4.6 & \textbf{0.1} \\
        PointMBF~\cite{yuan2023pointmbf} & 3DMatch & & 94.6 & 97.0 & 98.7 & 3.0 & \textbf{0.8} & \underline{81.0} & 92.0 & 97.1 & 6.2 & \textbf{2.1} & \underline{91.3} & 96.6 & 97.4 & 4.9 & \textbf{0.1} \\
         \ours{} (w/o SW, \emph{ours}) & 3DMatch & & \underline{95.5} & \underline{98.1} & \underline{99.6} & \underline{2.1} & \underline{0.9} & 78.3 & \underline{92.7} & \underline{98.0} & \underline{4.9} & 2.6 & 89.2 & \underline{97.2} & \underline{98.1} & \underline{3.6} & \underline{0.2} \\
         \ours{} (w/ SW, \emph{ours}) & 3DMatch & & \textbf{96.3} & \textbf{98.7} & \textbf{99.7} & \textbf{1.8} & \underline{0.9} & \textbf{81.9} & \textbf{94.7} & \textbf{98.5} & \textbf{4.2} & 2.4 & \textbf{91.4} & \textbf{97.9} & \textbf{98.6} & \textbf{3.1} & \textbf{0.1} \\
    \midrule
        UR\&R~\cite{el2021unsupervisedr} & ScanNet & & 92.7 & 95.8 & 98.5 & 3.4 & \underline{0.8} & 77.2 & 89.6 & 96.1 & 7.3 & 2.3 & 86.0 & 94.6 & 96.1 & 5.9 & \textbf{0.1} \\
        UR\&R (RGB-D) & ScanNet & & 94.1 & 97.0 & \underline{99.1} & 2.6 & \underline{0.8} & 78.4 & 91.1 & 97.3 & 5.9 & 2.3 & 87.3 & 95.6 & 97.2 & 5.0 & \textbf{0.1}\\
        BYOC~\cite{el2021bootstrap} & ScanNet & & 86.5 & 95.2 & \underline{99.1} & 3.8 & 1.7 & 56.4 & 80.6 & 96.3 & 8.7 & 4.3 & 78.1 & 93.9 & 96.4 & 5.6 & \underline{0.3} \\
        LLT~\cite{wang2022improving} & ScanNet & & 95.5 & 97.6 & \underline{99.1} & 2.5 & \underline{0.8} & 80.4 & 92.2 & 97.6 & 5.5 & 2.2 & 88.9 & 96.4 & 97.6 & 4.6 & \textbf{0.1} \\
        PointMBF~\cite{yuan2023pointmbf} & ScanNet & & 96.0 & 97.6 & 98.9 & 2.5 & \textbf{0.7} & \underline{83.9} & 93.8 & 97.7 & 5.6 & \textbf{1.9} & \underline{92.8} & 97.3 & 97.9 & 4.7 & \textbf{0.1} \\
        \ours{} (w/o SW, \emph{ours}) & ScanNet & & \underline{96.3} & \underline{98.7} & \textbf{99.8} & \underline{1.7} & 0.9 & 82.4 & \underline{94.7} & \underline{98.5} & \underline{4.1} & 2.3 & 91.4 & \underline{97.8} & \underline{98.6} & \underline{3.2} & \textbf{0.1}\\
        \ours{} (w/ SW, \emph{ours}) & ScanNet & & \textbf{97.6} & \textbf{99.1} & \textbf{99.8} & \textbf{1.4} & \underline{0.8} & \textbf{85.5} & \textbf{95.8} & \textbf{98.8} & \textbf{3.7} & \underline{2.1} & \textbf{93.1} & \textbf{98.4} & \textbf{98.9} & \textbf{2.9} & \textbf{0.1} \\

    \bottomrule
       
    \end{tabular}}
    \caption{\textbf{Evaluations on ScanNet under the 20-frame setting.} \textbf{Sup}: ground-truth pose supervision. \textbf{SW}: synthetic warming-up. $^{*}$ indicates that the models are trained with point cloud fragments instead of RGB-D sequences. \textbf{Boldfaced} numbers are the best and the second best are \underline{underlined}.}
    \label{tab:main_table_20frames}
\end{table*}

\para{Trained on ScanNet.}
For the $20$-frame setting, as shown in Tab.~\ref{tab:main_table_20frames}, although the results under this setting tend to be saturated, \ours{} still achieves consistent improvements on almost all metrics. Our method surpasses the previous state-of-the-art PointMBF by $1.6$ percent points on rotation accuracy under $5^{\circ}$, $1.6$ points on translation accuracy under $5$cm. And we also observe a significant decrease on the mean errors, with the relative improvements of $38\%$ to $44\%$. These results have shown that our method achieves consistent better registration results, which proves the effectiveness of our frame-to-model optimization framework.

\begin{table*}[t]
    \centering
        \small
     \setlength{\tabcolsep}{3.6pt}
    {\begin{tabular}{l|c|ccccc|ccccc|ccccc}
    \toprule
       \multicolumn{1}{l|}{\multirow{3}{*}{ }} & \multicolumn{1}{c|}{\multirow{3}{*}{Train Set}} & \multicolumn{5}{c|}{Rotation($^{\circ}$)} & \multicolumn{5}{c|}{Translation(cm)} & \multicolumn{5}{c}{Chamfer(mm)} \\
         &  & \multicolumn{3}{c}{Accuracy $\uparrow$} & \multicolumn{2}{c|}{Error$\downarrow$} &  \multicolumn{3}{c}{Accuracy $\uparrow$} & \multicolumn{2}{c|}{Error$\downarrow$} & \multicolumn{3}{c}{Accuracy $\uparrow$} & \multicolumn{2}{c}{Error$\downarrow$} \\
         &  & 5 & 10 & 45 & Mean & Med. & 5 & 10 & 25 & Mean & Med. & 1 & 5 & 10 & Mean & Med. \\
        
    \midrule
       UR$\&$R~\cite{el2021unsupervisedr} & 3DMatch & 40.4 & 51.0 & 73.0 & 36.4 & 9.3 & 22.6 & 34.6 & 50.6 & 76.2 & 24.0 & 29.4 & 42.6 
       & 48.5 & 179.7 & 11.7 \\
       LLT~\cite{wang2022improving} & 3DMatch & 50.4 & 60.0 & 77.1 & 30.6 & 4.8 & 30.2 & 44.2 & 59.0 & 58.0 & 13.9 & 38.4 & 51.8 & 57.3 & 165.8 & 3.8 \\
       PointMBF~\cite{yuan2023pointmbf} & 3DMatch & 59.3 & 62.5 & 76.6 & 25.5  & 3.2 & 34.2 & 47.9 & 61.6 & 51.2 & 8.2 & 42.9 & 55.8 & 60.2 & 103.1 & 1.3 \\
       \ours{} (w/o SW, \emph{ours}) & 3DMatch & \underline{66.0} & \underline{75.3} & \underline{86.2} & \underline{19.0} & \underline{2.6} & \underline{37.8} & \underline{58.6} & \underline{72.9} & \underline{40.9} & \underline{7.4} & \underline{49.7} & \underline{67.7} & \underline{71.9} & \underline{87.9} & \underline{1.0}\\
       \ours{} (w/ SW, \emph{ours}) & 3DMatch  & \textbf{72.6} & \textbf{81.1} & \textbf{91.1} & \textbf{12.5} & \textbf{2.2} & \textbf{44.6} & \textbf{65.9} & \textbf{78.5} & \textbf{28.5} & \textbf{5.8} & \textbf{56.1} & \textbf{73.6} & \textbf{77.1} & \textbf{72.0} & \textbf{0.7} \\
    \midrule
       UR$\&$R~\cite{el2021unsupervisedr} & ScanNet & 50.5 & 59.5 & 75.3 & 33.6 & 4.8 & 30.7 & 44.6 & 59.0 & 68.6 & 13.8 & 38.4 & 52.3 & 57.2 & 169.7 & 3.7\\
       LLT~\cite{wang2022improving} & ScanNet & 57.0 & 65.6 & 79.3 & 28.6 & 3.1 & 36.1 & 50.7 & 64.8 & 53.0 & 9.5 & 44.5 & 58.1 & 62.8 & 158.6 & 1.6\\
       PointMBF~\cite{yuan2023pointmbf} & ScanNet & 60.4 & 68.2 & 79.9 & 19.2 & 2.3 & 40.0 & 54.3 & 66.9 & 38.1 & 6.0 & 48.9 & 61.5 & 65.8 & 85.8 & 0.7 \\
        \ours{} (w/o SW, \emph{ours}) & ScanNet & \underline{74.4} & \underline{82.8} & \underline{92.3} & \textbf{10.8} & \underline{2.1} & \underline{46.8} & \underline{67.9} & \underline{80.4} & \textbf{25.4} & \underline{5.5} & \underline{58.5} & \underline{75.5} & \underline{79.0} & \underline{67.1} & \underline{0.6}\\
       \ours{} (w/ SW, \emph{ours}) & ScanNet & \textbf{77.4} & \textbf{84.5} & \textbf{92.5} & \underline{15.5} & \textbf{1.9} & \textbf{50.0} & \textbf{70.6} & \textbf{82.1} & \underline{30.1} & \textbf{5.0} & \textbf{61.5} & \textbf{77.6} & \textbf{80.9} & \textbf{73.8} & \textbf{0.5}  \\

    \bottomrule
    \end{tabular}}
    \caption{\textbf{Evaluations on ScanNet under the 50-frame setting.} \textbf{SW}: synthetic warming-up. \textbf{Boldfaced} numbers are the best and the second best are \underline{underlined}.}
    \label{tab:main_table_50frames}
\end{table*}

For the more challenging $50$-frame setting, our method achieves significant improvements over the baselines, as shown in Tab.~\ref{tab:main_table_50frames}.
\ours{}, without synthetic warming-up, surpasses PointMBF by over $14.0$, $6.8$ and $7.6$ points, respectively, on rotation, translation, and chamfer distance accuracies under the strictest thresholds, proving the superiority of our frame-to-model optimization framework.
Compared to the $20$-frame setting, the frame pairs in this setting have much smaller overlap with more severe lighting changes and geometric occlusion.
As the frame-to-frame optimization cannot provide effective training signal in these cases, the baselines thus fail to handle the inconsistency between two frames and therefore generate erroneous registrations.
On the contrary, thanks to the global encoding capability of the neural implicit field, our frame-to-model optimization can provide better training signal, which enables \ours{} to achieve better registration results in this setting.
Moreover, when the synthetic warming-up is adopted, we observe substantial improvements on almost all metrics. Notably, the accuracies under the strictest thresholds are further improved by $3.0$, $3.2$ and $3.0$ points, respectively.
These results are consistent with our motivation to provide stable high-quality initial poses for the neural implicit field. We would emphasize that, although the scenes in the Sim-RGBD dataset are simplistic and unrealistic, it has already provided effective initialization of the registration model.

\para{Trained on 3DMatch.}
We evaluate the generality of our method by training the models on 3DMatch. 
As shown in Tab.~\ref{tab:main_table_20frames}, we observe similar results to the models trained on ScanNet. Moreover, our method not only achieves new state-of-the-art results on nearly all metrics, especially on rotation accuracies, but also is on par with the baselines trained on ScanNet.
These results have demonstrated the strong generality of our method, even trained on a relatively small dataset.

For the $50$-frame setting, \ours{} also achieves more significant improvements over the baselines, as shown in Tab.~\ref{tab:main_table_50frames}. 
Our method, without synthetic warming-up, surpasses PointMBF by $6.7$ points on rotation accuracy under $5^{\circ}$, $3.6$ points on translation accuracy under $5$cm, $6.8$ points on chamfer distance accuracy under $1$mm.
And the model with synthetic warming-up further outperforms PointMBF by $13.3$ points on rotation accuracy under $5^{\circ}$, $10.4$ points on translation accuracy under $5$cm, $13.2$ points on chamfer distance accuracy under $1$mm. More notably, \ours{} without synthetic warming-up, trained on 3DMatch surpasses nearly all baselines trained on ScanNet, demonstrating the high data efficiency of our frame-to-model optimization strategy.
These results also suggest that, \emph{for unsupervised point cloud registration, a carefully designed training method is more important than scaling up the training data}.

\begin{figure*}[]
\centering
\includegraphics[width=\linewidth]{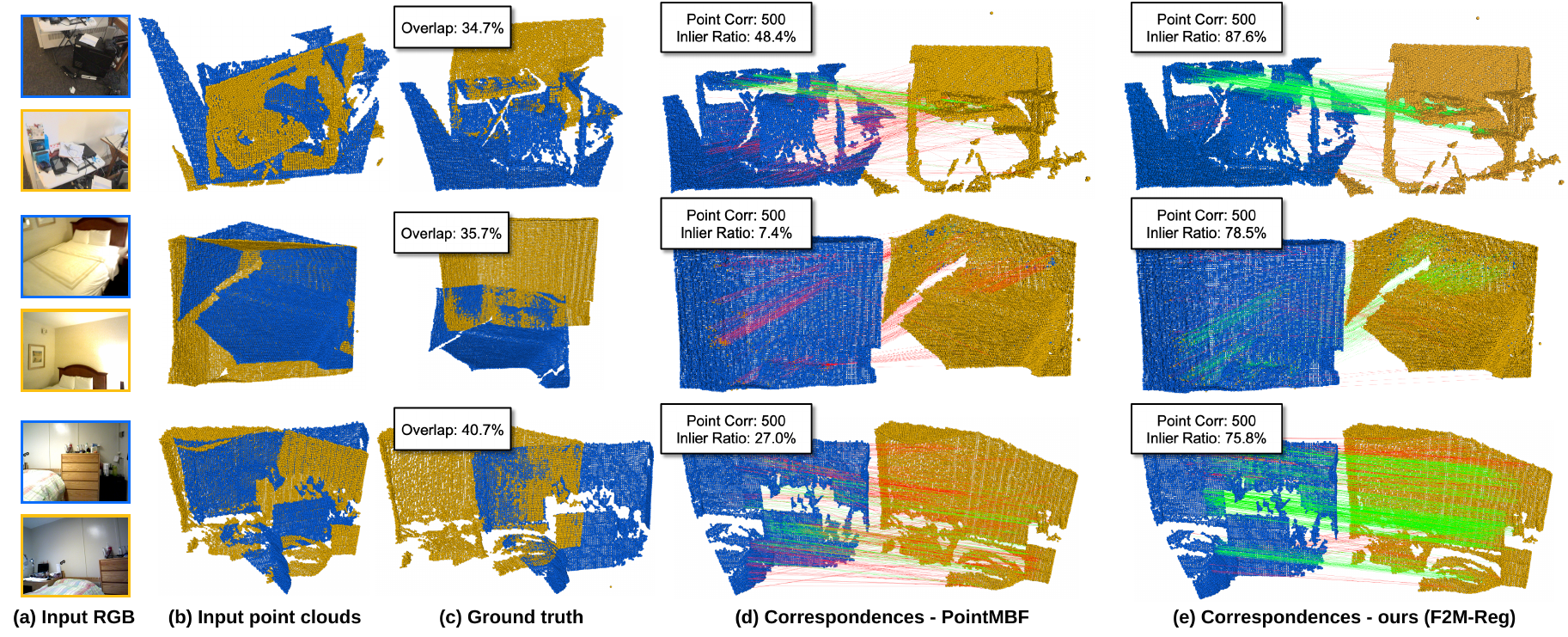}
\caption{\textbf{Correspondences of PointMBF and \ours{} on ScanNet and 3DMatch.} The first row shows that \ours{} outperforms when the input point clouds have a low overlap ratio. The subsequent rows illustrate that even under significant lighting changes, which adversely affect other methods, our approach continues to perform effectively.}
\label{fig:visual_result}
\end{figure*}
\para{Different overlaps and lighting changes.} 
We further investigate the effectiveness of our method under different overlaps and lighting changes in Tab.~\ref{tab:inlier_ratio}. 
For different overlaps, \ours{} significantly surpasses PointMBF when the overlap ratio between two frames is below $0.5$, demonstrating that our method can effectively handle the low-overlap scenarios.

For lighting changes, we use modified 3D-PSNR to measure lighting changes which computes the photometric differences between frames. Specifically, we first back-project RGB images to 3D point clouds and align the source point cloud to the target. Next, we select the top $30\%$ points in the overlapping area from the source point cloud with the largest RGB errors to their nearest neighbors in the target point cloud. This allows us to focus on the area with the most severe lighting changes. At last, we compute PSNR for these selected points. It is observed that our method achieves significant improvements on all PSNR intervals, especially when the PSNR is small, which proves the strong effectiveness of \ours{} under severe lighting changes.

\begin{table}[t]
    \centering
    \small
     \setlength{\tabcolsep}{1.5pt}
    {\begin{tabular}{l|c|cccc|cccc}
        \toprule
        & \multirow{2}{*}{Interval} & \multicolumn{4}{c|}{F2M-Reg} & \multicolumn{4}{c}{PointMBF} \\
        & & RRE & RTE & CD & IR 
        & RRE & RTE & CD & IR \\
        \midrule
        \multirow{4}{*}{Overlap} & $\leq 0.25$ & 47.7 & 87.5 & 185.9 & 19.8 & 86.1 & 162.8 & 263.9 & 17.5 \\
        & $(0.25, 0.5]$ & 9.1 & 18.3 & 21.9 & 61.1 & 20.9 & 39.5 & 40.3 & 52.6 \\
        & $(0.5, 0.75]$ & 3.6 & 9.3 & 8.7 & 72.7 & 4.4 & 10.6 & 7.8 & 65.1 \\
        & $> 0.75$ & 2.8 & 7.1 & 6.1 & 78.9 & 2.6 & 6.1 & 3.6 & 72.8 \\
        \midrule
        \multirow{4}{*}{3D-PSNR} & $\leq 2.0$  & 18.4 & 35.4 & 73.6 & 51.6 & 34.6 & 63.6 & 101.3 & 46.1 \\
        & $(2.0, 4.0]$ & 13.4 & 27.1 & 47.4 & 58.4 & 25.0 & 49.1 & 69.9 & 51.8 \\
        & $(4.0, 6.0]$ & 12.6 & 25.9 & 41.0 & 60.8 & 23.9 & 45.7 & 59.0 & 54.1 \\
        & $> 6.0$ & 14.1 & 25.7 & 39.9 & 62.7 & 24.7 & 48.8 & 59.4 & 55.1 \\
        \bottomrule
    \end{tabular}}
    \caption{\textbf{Evaluations under different overlaps and lighting changes on ScanNet.}  }
    \label{tab:inlier_ratio}
\end{table}

\para{Qualitative results.} Fig.~\ref{fig:visual_result} visualizes the correspondences from PointMBF~\cite{yuan2023pointmbf} and \ours{} on ScanNet and 3DMatch. 
Thanks to our frame-to-model optimization strategy, \ours{} extracts significantly more accurate correspondences in the scenes with smaller overlap ($1^{\text{st}}$ row) and severe lighting changes ($2^{\text{nd}}$ and $3^{\text{rd}}$ rows), which further proves the superiority of our design.

\begin{figure*}[t]
\centering
\includegraphics[width=\linewidth]{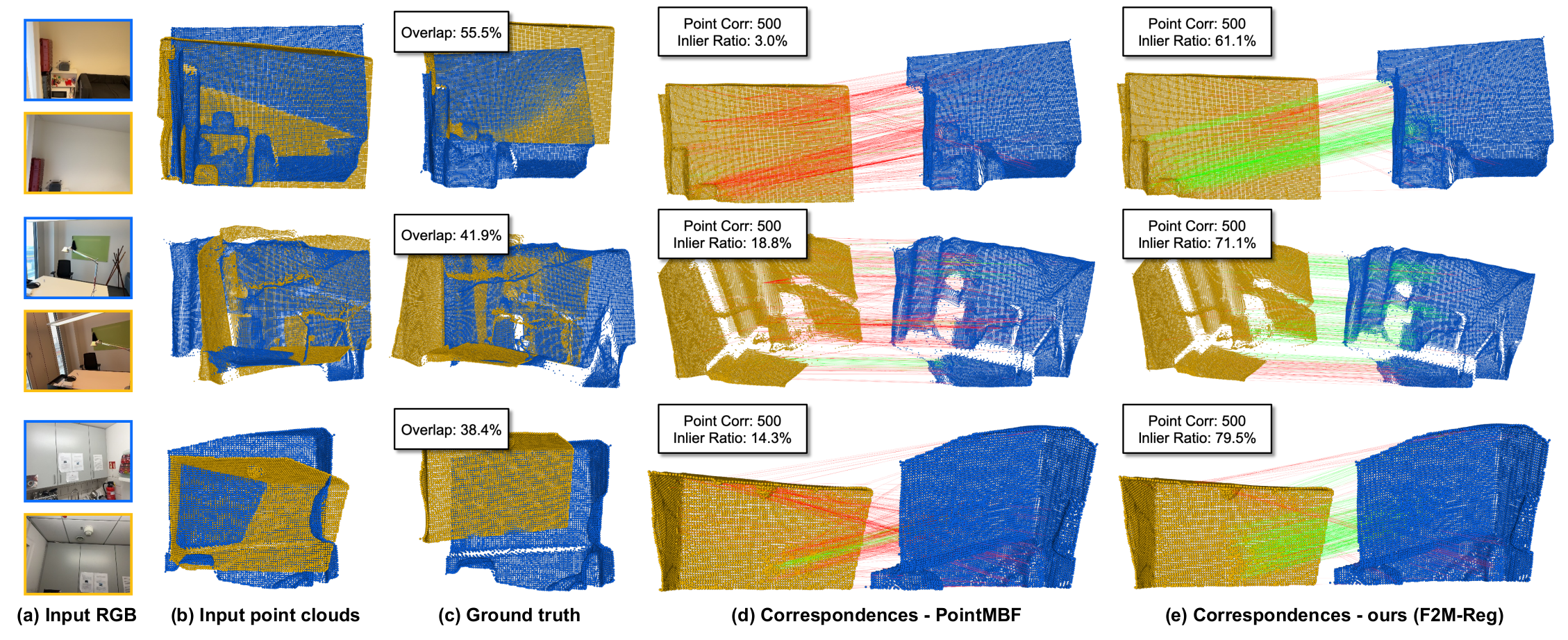}
\caption{\textbf{Comparison of the registration results on ScanNet++.} \ours{} demestrates strong recognition ability in complex scenes with drastic light changes and achieves higher inlier ratio.  }
\label{fig:visual_result_scannet++}
\end{figure*}
\begin{table*}
    \centering
    
        \small
     \setlength{\tabcolsep}{3.4pt}
    {\begin{tabular}{l|c|ccccc|ccccc|ccccc}
    \toprule
       \multicolumn{1}{c|}{\multirow{3}{*}{ }} & \multicolumn{1}{c|}{\multirow{3}{*}{Test Set}} & \multicolumn{5}{c|}{Rotation($^{\circ}$)} & \multicolumn{5}{c|}{Translation(cm)} & \multicolumn{5}{c}{Chamfer(mm)} \\
        &  & \multicolumn{3}{c}{Accuracy $\uparrow$} & \multicolumn{2}{c|}{Error$\downarrow$} &  \multicolumn{3}{c}{Accuracy $\uparrow$} & \multicolumn{2}{c|}{Error$\downarrow$} & \multicolumn{3}{c}{Accuracy $\uparrow$} & \multicolumn{2}{c}{Error$\downarrow$} \\
         &  & 5 & 10 & 45 & Mean & Med. & 5 & 10 & 25 & Mean & Med. & 1 & 5 & 10 & Mean & Med. \\
        
    \midrule
        UR\&R~\cite{el2021unsupervisedr} & 7-Scenes & 46.3 & 62.9 & 81.7 & 27.0 & 5.5 & 20.6 & 39.8 & 59.6 & 64.2 & 15.4 & 20.6 & 44.0 & 52.9 & 3747.8 & 7.4\\
        LLT~\cite{wang2022improving} & 7-Scenes & 53.8 & 69.4 & 83.5 & 22.5 & 4.4 & 19.6 & 43.1 & 64.8 & 44.4 & 12.1 & 20.8 & 48.1 & 56.2 & 1837.4 & 6.3\\
        PointMBF~\cite{yuan2023pointmbf} & 7-Scenes & 55.4 & 71.0 & 80.8 & 25.9 & 4.4 & 22.5 & 45.6 & 68.3 & 55.3 & 11.7 & 25.8 & \underline{51.2} & 58.5 & 1965.9 & 4.8\\
        \ours{} (w/o SW, \emph{ours}) & 7-Scenes & \underline{63.7} & \underline{81.0} & \underline{90.2} & \underline{15.3} & \underline{3.6} & \underline{24.2} & \underline{55.0} & \textbf{79.8} & \underline{58.4} & \underline{9.3} & \underline{27.7} & \textbf{62.7} & \underline{69.8} & \underline{1509.0} & \underline{2.9}\\
        \ours{} (w/ SW, \emph{ours}) & 7-Scenes & \textbf{65.6} & \textbf{81.7} & \textbf{91.9} & \textbf{13.2} & \textbf{3.3} & \textbf{30.4} & \textbf{57.7} & \underline{79.2} & \textbf{24.6} & \textbf{7.6} & \textbf{32.1} & \textbf{62.7} & \textbf{70.2} & \textbf{1631.2} & \textbf{2.3}\\
    \midrule
        UR\&R~\cite{el2021unsupervisedr} & ScanNet++ & 56.4 & 64.8 & 77.7 & 30.4 & 3.4 & 32.9 & 48.5 & 62.9 & 407.3 & 10.8 & 44.2 & 59.1 & 63.6 & 145.8 & 1.7\\
        LLT~\cite{wang2022improving} & ScanNet++ & 64.4 & 71.8 & 81.7 & 25.3 & 2.3 & 40.2 & 56.8 & 70.4 & 387.9 & 7.4 & 51.5 & 66.3 & 70.3 & 131.5 & 0.9\\
        PointMBF~\cite{yuan2023pointmbf} & ScanNet++ & 60.8 & 67.3 & 77.9 & 30.1 & 2.4 & 41.1 & 55.2 & 66.4 & \underline{400.2} & 7.4 & 50.7 & 62.8 & 66.2 & 143.0 & 0.9\\
        \ours{} (w/o SW, \emph{ours}) & ScanNet++ & \underline{72.5} & \underline{81.4} & \underline{87.8} & \underline{11.3} & \underline{2.2} & \underline{43.4} & \underline{65.3} & \underline{79.4} & 804.5 & \underline{6.0} & \underline{57.3} & \underline{74.8} & \underline{78.6} & \underline{123.6} & \underline{0.7}\\
        \ours{} (w/ SW, \emph{ours}) & ScanNet++ & \textbf{75.7} & \textbf{82.5} & \textbf{90.1} & \textbf{14.1} & \textbf{1.7} & \textbf{50.4} & \textbf{69.7} & \textbf{80.5} & \textbf{371.4} & \textbf{4.9} & \textbf{61.6} & \textbf{77.1} & \textbf{80.0} & \textbf{107.0} & \textbf{0.5}\\
    \bottomrule
    \end{tabular}}
    \caption{\textbf{Evaluations on 7-Scenes and ScanNet++ under the 50-frame setting.} \textbf{SW}: synthetic warming-up. \textbf{Boldfaced} numbers are the best and the second best are \underline{underlined}.}
    \label{tab:Comparison_tum_rgbd_ScanNet++}
\end{table*}

\subsection{Evaluations on 7-Scenes and ScanNet++}
\label{subsec: ex_res_tum}

\para{Dataset.} 7-Scenes~\cite{shotton2013scene} and ScanNet++~\cite{Yeshwanth_2023_ICCV} are two more accurate and high-resolution RGB-D datasets. 7-Scenes is a dataset with slow camera motion, which is often used in the evaluation of Simultaneous localization and mapping (SLAM) tasks. ScanNet++ is more often used for novel view synthesis due to its high resolution images and faster camera movement. We conduct testing on these two dataset, where 7 scenes are used on 7-Scenes and 336 scenes are used on ScanNet++. The data used on ScanNet++ is from the iPhone RGB-D sequences. All registration models in Tab.~\ref{tab:Comparison_tum_rgbd_ScanNet++} are trained on ScanNet and tested under 50 frames apart.

\para{Tested on 7-Scenes.}
As shown in Tab.~\ref{tab:Comparison_tum_rgbd_ScanNet++}~(top), compared to the previous state-of-the-art PointMBF, \ours{} without synthetic warming-up achieves the improvements of $8.3$ points on rotation accuracy under $5^{\circ}$, $1.7$ points on translation accuracy under $5$cm, and $4.4$ points on chamfer distance accuracy under $1$mm. And \ours{} with synthetic warming-up further improves the performance by $1.9$ points on rotation accuracy under $5^{\circ}$, $6.2$ points on translation accuracy under $5$cm, and $4.4$ points on chamfer distance accuracy under $1$mm.
These results have proven the effectiveness of our method on dataset with slow camera motion and the cross-dataset generality of our method.

\para{Tested on ScanNet++.}
As shown in Tab.~\ref{tab:Comparison_tum_rgbd_ScanNet++}~(bottom), our method without synthetic warming-up surpasses PointMBF by $11.7$ points on rotation accuracy under $5^{\circ}$, $2.3$ points on translation accuracy under $5$cm, and $6.6$ points on chamfer distance accuracy under $1$mm.
And the improvements increase to $14.9$, $9.3$ and $10.9$ points on the three metrics, respectively, when synthetic warming-up is adopted.
Note that our \ours{} achieves more considerable improvements on ScanNet++ than 7-Scenes, further demonstrating the superiority of our method in fast-motion scenarios where the inconsistent factors between frames are more significant.

\para{Qualitative results.}
Fig.~\ref{fig:visual_result_scannet++} visualizes the correspondences extracted by PointMBF and \ours{} on ScanNet++.
Our method can successfully handle these hard cases with obvious color changes and relatively low overlap, while PointMBF struggles to extract reasonable correspondences. These visualizations further proves the effectiveness of our frame-to-model optimization design.

\begin{table*}[]
    \centering
    \small
     \setlength{\tabcolsep}{4.2pt}
    {\begin{tabular}{l|ccccc|ccccc|cccccc}
    \toprule
        \multicolumn{1}{c|}{\multirow{3}{*}{Model}} & \multicolumn{5}{c|}{Rotation($^{\circ}$)} & \multicolumn{5}{c|}{Translation(cm)} & \multicolumn{5}{c}{Chamfer(mm)} \\
         &  \multicolumn{3}{c}{Accuracy $\uparrow$} & \multicolumn{2}{c|}{Error$\downarrow$} &  \multicolumn{3}{c}{Accuracy $\uparrow$} & \multicolumn{2}{c|}{Error$\downarrow$} & \multicolumn{3}{c}{Accuracy $\uparrow$} & \multicolumn{2}{c}{Error$\downarrow$} \\
          & 5 & 10 & 45 & Mean & Med. & 5 & 10 & 25 & Mean & Med. & 1 & 5 & 10 & Mean & Med. \\
        
    \midrule
         (a.1) frame-to-frame & 74.2 & 81.8 & 89.0 & \textbf{15.5} & 2.0 & 46.3 & 67.8 & 79.8 & 31.9 & 5.6 & 58.4 & 75.4 & 78.9 & 72.7 & 0.6\\
         (a.2) frame-to-model$^{*}$ & \textbf{77.4} & \textbf{84.5} & \textbf{92.5} & \textbf{15.5} & \textbf{1.9} & \textbf{50.0} & \textbf{70.6} & \textbf{82.1} & \textbf{30.1} & \textbf{5.0} & \textbf{61.5} & \textbf{77.6} & \textbf{80.9} & \textbf{73.8} & \textbf{0.5}\\
    \midrule
        (b.1) warm up only & 71.3 & 78.6 & 87.4 & 15.8 & \underline{2.0} & 46.9 & 65.5 & 76.3 & 34.5 & 5.4 & 57.5 & 72.2 & 75.0 & 77.7 & \underline{0.6} \\
        (b.2) w/o pose optimization & 76.0 & 84.3 & \underline{92.7} & \underline{10.3} & \underline{2.0} & 46.8 & 69.4 & \underline{81.8} & \underline{23.8} & 5.4 & 59.4 & \underline{77.1} & \underline{80.6} & \underline{61.5} & \underline{0.6} \\
        (b.3) w/ 20 tracking iterations & \underline{76.7} & \underline{84.4} & \textbf{92.9} & \textbf{10.0} & \textbf{1.9} & \underline{48.4} & \underline{69.7} & \underline{81.8} & \textbf{22.7} & \underline{5.2} & \underline{60.7} & \underline{77.1} & 80.2 & \textbf{60.4} & \underline{0.6} \\
         (b.4) w/ 100 tracking iterations$^{*}$ & \textbf{77.4} & \textbf{84.5} & 92.5 & 15.5 & \textbf{1.9} & \textbf{50.0} & \textbf{70.6} & \textbf{82.1} & 30.1 & \textbf{5.0} & \textbf{61.5} & \textbf{77.6} & \textbf{80.9} & 73.8 & \textbf{0.5}\\
    \midrule
        (c.1) warm up only & 71.3 & 78.6 & 87.4 & 15.8 & 2.0 & 46.9 & 65.5 & 76.3 & 34.5 & 5.4 & 57.5 & 72.2 & 75.0 & 77.7 & \underline{0.6} \\
        (c.2) w/o circle loss & 71.0 & 77.7 & 86.7 & 18.8 & 2.0 & 48.7 & 66.0 & 75.8 & 34.7 & 5.2 & 58.1 & 72.2 & 75.0 & 83.8 & \underline{0.6} \\
        (c.3) w/o correspondence loss & \underline{77.1} & \underline{84.3} & \textbf{92.7} & \textbf{10.3} & \textbf{1.9} & \underline{49.1} & \underline{70.2} & \underline{81.6} & \textbf{24.0} & \underline{5.1} & \underline{61.1} & \underline{77.3} & \underline{80.4} & \textbf{72.7} & \textbf{0.5} \\
        (c.4) full* & \textbf{77.4} & \textbf{84.5} & \underline{92.5} & \underline{15.5} & \textbf{1.9} & \textbf{50.0} & \textbf{70.6} & \textbf{82.1} & \underline{30.1} & \textbf{5.0} & \textbf{61.5} & \textbf{77.6} & \textbf{80.9} & \underline{73.8} & \textbf{0.5}\\

    \bottomrule
    \end{tabular}}
     \caption{\textbf{Ablation studies on ScanNet.} * indicates the default settings of \ours. \textbf{Boldfaced} numbers are the best and the second best are \underline{underlined}.}
    \label{tab:Abl_fitted_version}
\end{table*}

\subsection{Ablation Study}
\label{subsec: abl_study}

We conduct comprehensive ablation studies on ScanNet to investigate the contributions of different components in our pipeline. Unless otherwise mentioned, the models use the same checkpoint warmed up on Sim-RGBD in the experiments.

\para{Frame-to-model optimization.}
To study the effectiveness of our frame-to-model optimization, we first compare two optimization methods in Tab.~\ref{tab:Abl_fitted_version} (a): (1) frame-to-frame optimization~\cite{el2021unsupervisedr}, and (2) frame-to-model optimization.
The frame-to-model optimization consistently achieves significant improvements over the frame-to-frame optimization on all metrics, which further demonstrates the effectiveness of our design. By encoding the radiance and the geometric information of the scene with a neural implicit field, our method provides more effective training signal for the registration model, which leads to better registration performance.

\begin{table}[t]
    \centering
    \small
    \setlength{\tabcolsep}{10.0pt}
    \begin{tabular}{l|cc|cc}
    \toprule
        & \multicolumn{2}{c|}{Rotation ($^{\circ}$)} & \multicolumn{2}{c}{Translation (cm)} \\
        & mean & median & mean & median \\
    \midrule
        frame-to-frame & 4.9 & 2.2 & 13.0 & 6.0\\
        frame-to-model & \textbf{2.9} & \textbf{0.8} & \textbf{7.0} & \textbf{3.0}\\
    \bottomrule
    \end{tabular}
     \caption{\textbf{Ablation study of pose quality on ScanNet \texttt{scene0000\_00}.} The results are statistics of the pose variations for pairs with rotation angles greater than $20^{\circ}$. \textbf{Boldfaced} numbers are the best.}
    \label{tab:pose_optimization_scannet}
\end{table}

To compare the pose quality obtained from the two optimization methods during training, we evaluated the optimized poses of the two methods during the first training epoch in Tab.~\ref{tab:pose_optimization_scannet}.
The results are computed on the pairs with rotations greater than $20^{\circ}$ in \texttt{scene0000\_00} of ScanNet.
Our frame-to-model optimization achieves significantly better poses than the frame-to-frame optimization during training, with the relative improvements of over $40\%$ on mean rotation error, $63\%$ on median rotation error, $46\%$ on mean translation error, and $50\%$ on median translation error.
These improvements in pose estimation during training provide stronger supervision signals for the registration model.

At last, we investigate the detailed configurations of the frame-to-model optimization in Tab.~\ref{tab:Abl_fitted_version} (b). 
We compare four models: (1) the model obtained from synthetic warming-up, (2) the model without pose optimization, (3) the model with $20$ tracking iterations, and (4) the model with $100$ tracking iterations.
For the model (2), the estimated initial pose is directly used to train the registration model.
From the results, we have three important findings: First, \emph{fine-tuning the registration model on real-world data is critical}, as there is a dramatic gap between the distribution between synthetic and real-world data. Second, \emph{generating high-quality poses are critical to provide effective training signals}, as demonstrated by the consistent improvements when more tracking iterations are used.

\begin{figure}[]
\centering
\includegraphics[width=\linewidth]{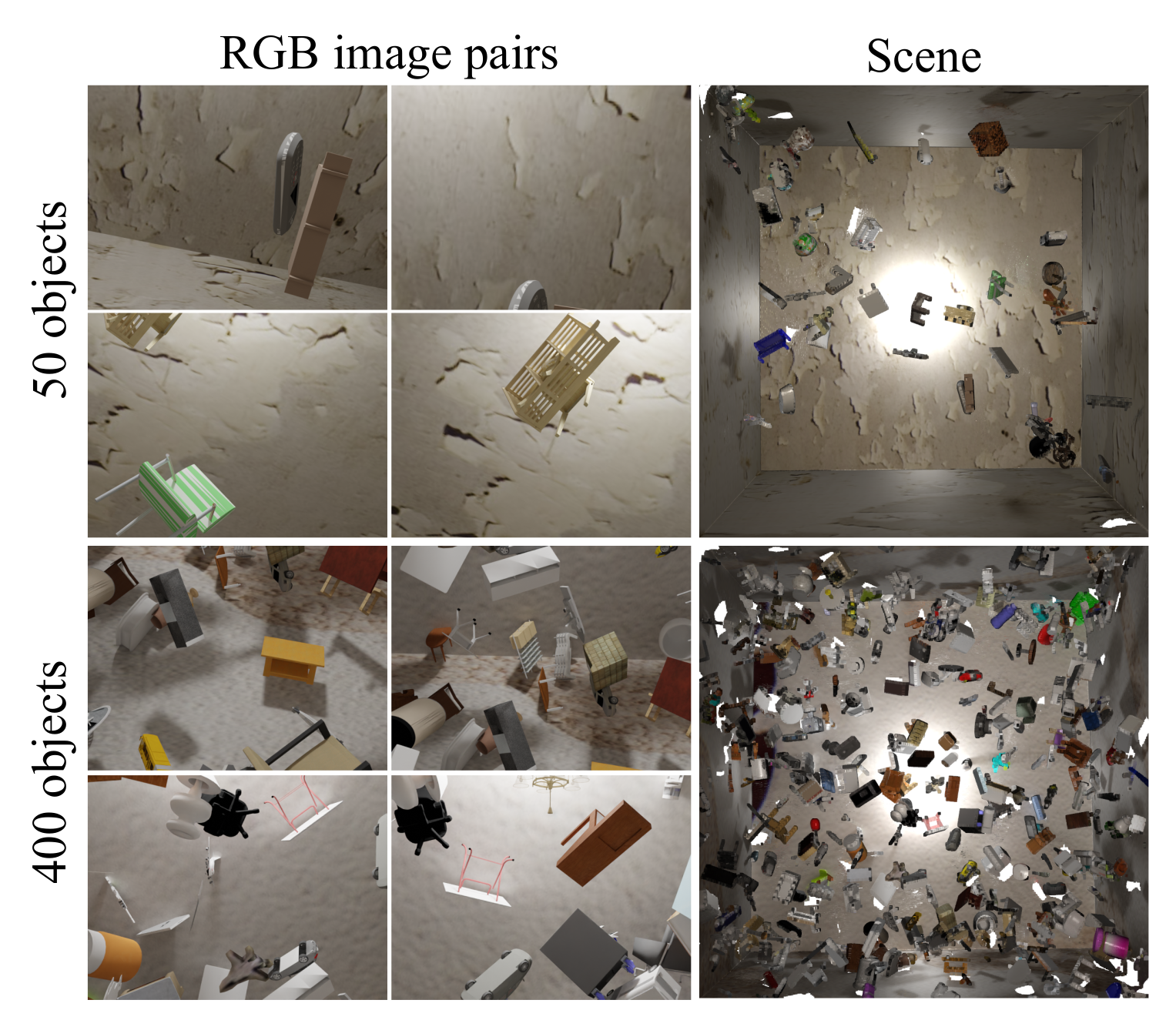}
\caption{\textbf{Visualization on 50 objects vs 400 objects.} By row, the top row corresponds to a scene with 50 objects, while the bottom row represents a scene with 400 objects. By column, the RGB image pairs depict the image pairs used for training in the scene, while the scene column illustrates the overall style of the scene.}
\label{fig:visual_num_obj}
\end{figure}

\begin{table*}[]
    \centering
    \small
    \setlength{\tabcolsep}{5.9pt}
    {\begin{tabular}{l|ccccc|ccccc|ccccc}
    \toprule
        \multicolumn{1}{c|}{\multirow{3}{*}{Model}} & \multicolumn{5}{c|}{Rotation($^{\circ}$)} & \multicolumn{5}{c|}{Translation(cm)} & \multicolumn{5}{c}{Chamfer(mm)} \\
         &  \multicolumn{3}{c}{Accuracy $\uparrow$} & \multicolumn{2}{c|}{Error$\downarrow$} &  \multicolumn{3}{c}{Accuracy $\uparrow$} & \multicolumn{2}{c|}{Error$\downarrow$} & \multicolumn{3}{c}{Accuracy $\uparrow$} & \multicolumn{2}{c}{Error$\downarrow$} \\
         & 5 & 10 & 45 & Mean & Med. & 5 & 10 & 25 & Mean & Med. & 1 & 5 & 10 & Mean & Med. \\
    \midrule
         (a.1) 50 objects & \textbf{72.3} & \textbf{80.3} & \textbf{88.9} & \textbf{15.6} & 2.2 & 43.2 & 64.5 & \textbf{78.3} & \textbf{32.3} & 6.1 & 55.1 & \textbf{73.1} & \textbf{77.0} & \textbf{72.6} & 0.7\\
         (a.2) 400 objects$^{*}$ & 71.3 & 78.6 & 87.4 & 15.8 & \textbf{2.0} & \textbf{46.9} & \textbf{65.5} & 76.3 & 34.5 & \textbf{5.4} & \textbf{57.5} & 72.2 & 75.0 & 77.7 & \textbf{0.6} \\
    \midrule
         (b.1) 20 scenes* & 71.3 & 78.6 & 87.4 & 15.8 & \underline{2.0} & 46.9 & 65.5 & 76.3 & 34.5 & \textbf{5.4} & 57.5 & 72.2 & 75.0 & 77.7 & \textbf{0.6}\\
         (b.2) 40 scenes & 72.5 & \textbf{80.3} & \textbf{89.6} & \underline{14.6} & 2.1 & 46.3 & 66.2 & \underline{77.8} & 31.9 & \underline{5.6} & 57.0 & \textbf{73.8} & \underline{76.9} & \textbf{72.4} & \textbf{0.6}\\
         (b.3) 60 scenes & \underline{72.6} & 79.5 & \underline{88.2} & 14.9 & 2.1 & \underline{48.1} & \underline{66.5} & 77.1 & \underline{31.0} & \underline{5.6} & \underline{58.4} & 73.2 & 76.1 & 73.6 & \textbf{0.6}\\
         (b.4) 80 scenes & \textbf{73.5} & \underline{80.1} & 88.1 & \textbf{14.1} & \textbf{1.9} & \textbf{48.7} & \textbf{67.3} & \textbf{78.0} & \textbf{30.2} & \textbf{5.4} & \textbf{58.9} & \textbf{74.0} & \textbf{77.0} & \underline{73.4} & \textbf{0.6}\\
    \bottomrule
    \end{tabular}}
     \caption{\textbf{Ablation studies of synthetic dataset on ScanNet.} * indicates the default settings of \ours. \textbf{Boldfaced} numbers are the best and the second best are \underline{underlined}.}
    \label{tab:Abl_synthetic}
\end{table*}

\begin{table*}[]
    \centering
    \small
    \setlength{\tabcolsep}{4.3pt}
    {\begin{tabular}{l|ccccc|ccccc|ccccc}
    \toprule
        \multicolumn{1}{c|}{\multirow{3}{*}{Number of Scenes}} & \multicolumn{5}{c|}{Rotation($^{\circ}$)} & \multicolumn{5}{c|}{Translation(cm)} & \multicolumn{5}{c}{Chamfer(mm)} \\
        &  \multicolumn{3}{c}{Accuracy $\uparrow$} & \multicolumn{2}{c|}{Error$\downarrow$} &  \multicolumn{3}{c}{Accuracy $\uparrow$} & \multicolumn{2}{c|}{Error$\downarrow$} & \multicolumn{3}{c}{Accuracy $\uparrow$} & \multicolumn{2}{c}{Error$\downarrow$} \\
        & 5 & 10 & 45 & Mean & Med. & 5 & 10 & 25 & Mean & Med. & 1 & 5 & 10 & Mean & Med. \\
        
    \midrule
         300 (ROSEFusion~\cite{zhang2021rosefusion})  & 70.5 & 79.6 & 90.4 & 12.5 & 2.3 & 42.9 & 64.5 & 77.4 & 28.7 & 5.9 & 55.0 & 72.2 & 75.7 & 73.1 & \underline{0.7}\\
         +200 (ROSEFusion~\cite{zhang2021rosefusion}+Ours) & 71.1 & 80.6 & 91.2 & 12.0 & 2.3 & 42.5 & 64.4 & 77.8 & 27.5 & 6.0 & 54.7 & 72.5 & 76.1 & 70.8 & \underline{0.7} \\
         +400 (ROSEFusion~\cite{zhang2021rosefusion}+Ours) & \underline{71.6} & 80.6 & 90.9 & 11.4 & \underline{2.2} & 43.3 & 64.7 & 77.7 & 27.0 & \underline{5.9} & 55.2 & 72.7 & 76.2 & 69.2 & \underline{0.7} \\
         +600 (ROSEFusion~\cite{zhang2021rosefusion}+Ours) & \textbf{73.2} & \underline{81.9} & \underline{91.6} & \underline{11.1} & \textbf{2.1} & \textbf{45.8} & \underline{66.5} & \textbf{79.7}  & \underline{25.6} & \textbf{5.5} & \textbf{57.4} & \underline{74.4} & \textbf{78.1} & \underline{65.9} & \textbf{0.6} \\
         +745 (ROSEFusion~\cite{zhang2021rosefusion}+Ours) & \textbf{73.2} & \textbf{82.1} & \textbf{91.7} & \textbf{10.7} & \textbf{2.1} & \underline{45.3} & \textbf{66.6} & \underline{79.6} & \textbf{24.7} & \textbf{5.5} & \underline{57.1} & \textbf{74.5} & \underline{78.0} & \textbf{63.0} & \textbf{0.6} \\
    \bottomrule
    \end{tabular}}
     \caption{\textbf{Comparisons to SLAM on ScanNet.} \textbf{Boldfaced} numbers are the best and the second best are \underline{underlined}. }
    \label{tab:Abl_SLAM}
\end{table*}

\begin{table*}[h]
    \centering
    \small
    \setlength{\tabcolsep}{6.9pt}
    {\begin{tabular}{l|ccccc|ccccc|ccccc}
    \toprule
        \multicolumn{1}{l|}{\multirow{3}{*}{\# Scenes}} & \multicolumn{5}{c|}{Rotation($^{\circ}$)} & \multicolumn{5}{c|}{Translation(cm)} & \multicolumn{5}{c}{Chamfer(mm)} \\
         & \multicolumn{3}{c}{Accuracy $\uparrow$} & \multicolumn{2}{c|}{Error$\downarrow$} &  \multicolumn{3}{c}{Accuracy $\uparrow$} & \multicolumn{2}{c|}{Error$\downarrow$} & \multicolumn{3}{c}{Accuracy $\uparrow$} & \multicolumn{2}{c}{Error$\downarrow$} \\
         & 5 & 10 & 45 & Mean & Med. & 5 & 10 & 25 & Mean & Med. & 1 & 5 & 10 & Mean & Med. \\
        
    \midrule
         200  & 71.6 & 79.5 & 88.0 & 16.4 & 2.1 & 45.8 & 65.5 & 77.7 & 32.3 & 5.6 & 57.0 & 73.0 & 76.3 & 77.4 & 0.7\\
         400 & 71.9 & 79.1 & 88.3 & 16.7 & \underline{2.0} & 47.4 & 66.4 & 77.1 & 31.2 & 5.4 & 58.3 & 73.1 & 76.1 & 77.3 & \underline{0.6}\\
         600 & 72.1 & 79.4 & 88.0 & 16.4 & \underline{2.0} & 48.2 & 66.6 & 77.5 & 31.8 & 5.3 & 58.5 & 73.2 & 76.2 & 78.3 & \underline{0.6} \\
         800 & \underline{74.4} & \underline{81.4} & \underline{89.2} & \textbf{14.7} & \underline{2.0} & \underline{48.9} & \underline{68.4} & \underline{79.5}  & \textbf{28.8} & \underline{5.2} & \underline{59.6} & \underline{75.4} & \underline{78.4} & \textbf{70.1} & \underline{0.6} \\
         1045 & \textbf{77.4} & \textbf{84.5} & \textbf{92.5} & \underline{15.5} & \textbf{1.9} & \textbf{50.0} & \textbf{70.6} & \textbf{82.1} & \underline{30.1} & \textbf{5.0} & \textbf{61.5} & \textbf{77.6} & \textbf{80.9} & \underline{73.8} & \textbf{0.5} \\
    \bottomrule
    \end{tabular}}
     \caption{\textbf{Ablation of data scability on ScanNet.} \textbf{Boldfaced} numbers are the best and the second best are \underline{underlined}. }
    \label{tab:Abl_scability}
\end{table*}

\para{Loss functions.}
We further ablate the loss functions used in our method in Tab.~\ref{tab:Abl_fitted_version} (c).
Circle loss is the most important loss function as it directly supervises the point features from the registration model and ablating the circle loss leads to severe performance drop. The correspondence loss slightly increases the performance. The combination of the two loss functions achieves the best registration results.

\para{Synthetic dataset.}
Next, we study how the construction of the synthetic dataset affects the registration model. For simplicity, we directly compare the model warmed up on the synthetic dataset.
We first investigate the influence of the number of objects in each scene.
To this end, we compare the models warmed up on two datasets in Tab.~\ref{tab:Abl_synthetic} (a), i.e., a sparse one with $50$ objects per scene and a dense one with $400$ objects per scene.
Fig.~\ref{fig:visual_num_obj} visualizes the comparison of the two datasets. The model warmed up on the sparse dataset performs comparably to the model warmed up on the sparse dataset. This demonstrates that the synthetic warming-up is not sensitive to the object number in the scene as long as the scene can provide sufficient geometric information.

Next, we investigate the influence of the size of dataset by comparing the model warmed up on $20$, $40$, $60$ and $80$ scenes, respectively.
As shown in Tab.~\ref{tab:Abl_synthetic} (b), increasing the dataset size effectively improves performance, but the gains become less significant when the dataset size increases from $60$ to $80$. This suggests that the synthetic warming-up does not rely on a large synthetic dataset.
Actually, as a low-level vision task, point cloud registration is less sensitive to the dataset size than high-level vision tasks such as object detection and segmentation, so we can effectively warm up the registration with a small synthetic dataset.

\para{SLAM v.s. neural implicit field.}
Given RGB-D sequences, another popular method to estimate the frame poses to leverage a SLAM system, and the poses can be then used to train the registration model.
However, the poses generated by SLAM are usually not accurate enough to effectively train the registration model, while our frame-to-model optimization can further enhance the performance of the registration model.
To verify this, we first conduct SLAM on $300$ scenes to estimate the frame poses and train the registration model with the estimated poses in a supervised manner, similar to our synthetic warming-up.
Afterwards, we train the resultant registration model on the other $745$ scenes with the frame-to-model optimization.
As shown in Tab.~\ref{tab:Abl_SLAM}, our method consistently improves the performance of the model as more scenes are used in the frame-to-model optimization.
Notably, the model initialized with ROSEFusion performs worse than the model with our synthetic warming-up, which further proves the effectiveness of our design.

\para{Data scalability.}
\ours{} follows a scene-by-scene training paradigm, which allows it to sequentially include more scenes during training and persistently improves the performance of the registration model. We evaluate this property in Tab.~\ref{tab:Abl_scability}, where we progressively increase the training scenes from $200$ to $1045$. The registration performance improves consistently as more scenes are incorporated. Specifically, when the number of scenes increases from $200$ to $1045$, the rotation accuracy under $5^{\circ}$ improves by $8$ points, the translation accuracy under $5$cm improves by $4.2$ points, and the chamfer distance accuracy under $1$mm improves by $4.5$ points. The property indicates that our method can benefit from a training paradigm similar to online learning or lifelong learning, where our registration model could gain continual performance improvements as more data are captured. We think this could potentially help construct a foundation model for 3D registration tasks.

%%%%%%%%%%%%%%%%% CONCLUSION
% !Tex root = main.tex

\section{Conclusion}

We present \ours{}, a frame-to-model optimization framework for unsupervised RGB-D point cloud registration. By leveraging the neural implicit field as a global scene model, our method captures the consistency across multiple views re-rendered from the global model, enabling more accurate pose optimization. This design significantly enhances robustness against lighting changes, geometric occlusions, and reflective materials. Additionally, We implement a warming-up mechanism on a synthetic dataset to initialize neural field optimization effectively. Extensive experiments on four datasets demonstrate effectiveness of our methods.

\para{Limitations.} Despite achieving state-of-the-art performance, our method has some limitations. First, applying it to outdoor scenes is still challenging. This is mainly because constructing neural implicit fields for outdoor scenes encounters difficulties such as excessive depth ranges and limited model capacity. Second, as our method optimizes a neural implicit field on the fly, the training time is relatively long.

\para{Future work.} In addition to the aforementioned limitations, there are several promising directions for extending \ours{}, such as dynamic scene registration and non-rigid registration. And our method has shown strong data scalability and thus could contribute to the construction of foundation model for point cloud registration. We would like to explore these topics in the future.

\bibliography{main}
\bibliographystyle{IEEEtran}

\newpage

\begin{IEEEbiography}[{\includegraphics[width=1in,height=1.25in,clip,keepaspectratio]{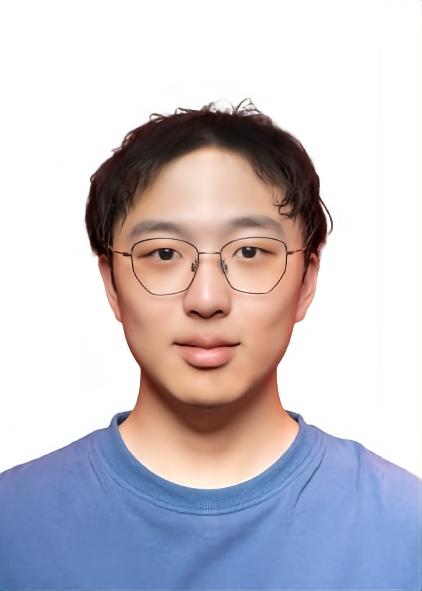}}]{Zhinan Yu} received his B.E. degree in computer science and technology from Northwest Agriculture and Forest University (NWAFU) in 2022. He is now pursuing the Ph.D. degree at the National University of Defense Technology (NUDT). His research interests focus on 3D vision and robotic perception.
\end{IEEEbiography}

\begin{IEEEbiography}[{\includegraphics[width=1in,height=1.25in,clip,keepaspectratio]{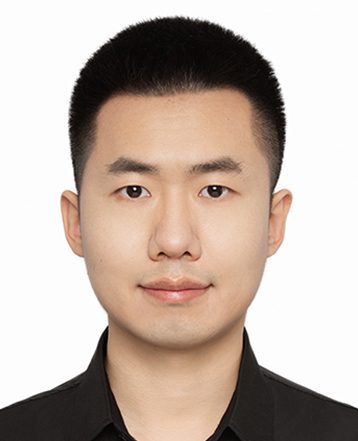}}]{Zheng Qin} received the B.E., M.E. and Ph.D. degrees in computer science from National University of Defense Technology (NUDT), China, in 2016, 2018 and 2023, respectively. He is currently an assistant professor at Defense Innovation Institute, China. His research interests focus on 3D vision, computer graphics and embodied perception.
\end{IEEEbiography}

\begin{IEEEbiography}[{\includegraphics[width=1in,height=1.25in,clip,keepaspectratio]{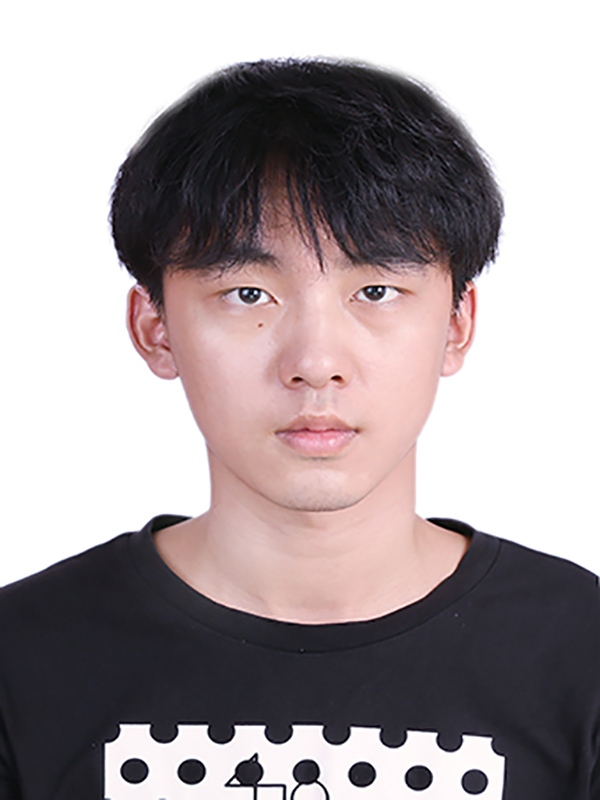}}]{Yijie Tang}
Yijie Tang is currently pursuing the Ph.D. degree at the National University of Defense Technology (NUDT). Before that he got the bachelor’s degree from Wuhan University and received the master’s degree from the National University of Defense Technology (NUDT). His current research interests include 3D reconstruction, SLAM, and embodied perception.
\end{IEEEbiography}

\begin{IEEEbiography}[{\includegraphics[width=1in,height=1.25in,clip,keepaspectratio]{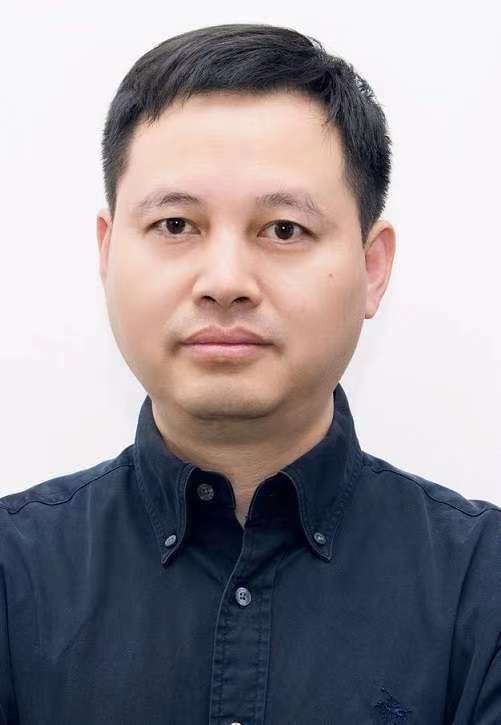}}]{Yongjun Wang}
received the Ph.D. degree in computer architecture from the National University of Defense Technology, China, in 1998. He is currently as a Full Professor with the College of Computer, National University of Defense Technology, Changsha, China. His research interests include artificial intelligence, network security and system security. 
\end{IEEEbiography}

\begin{IEEEbiography}[{\includegraphics[width=1in,height=1.25in,clip,keepaspectratio]{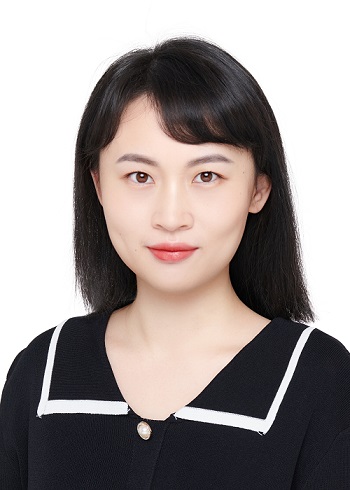}}]{Renjiao Yi} is an Associate Professor at the School of Computer Science, National University of Defense Technology. She received a Bachelor’s degree in Computer Science from National University of Defense Technology, China, in 2013, and a Ph.D. from Simon Fraser University, Burnaby, BC, Canada, in 2019. Her research interests include inverse rendering, image relighting, scene reconstruction in 3D vision and graphics. 
\end{IEEEbiography}

\begin{IEEEbiography}[{\includegraphics[width=1in,height=1.25in,clip,keepaspectratio]{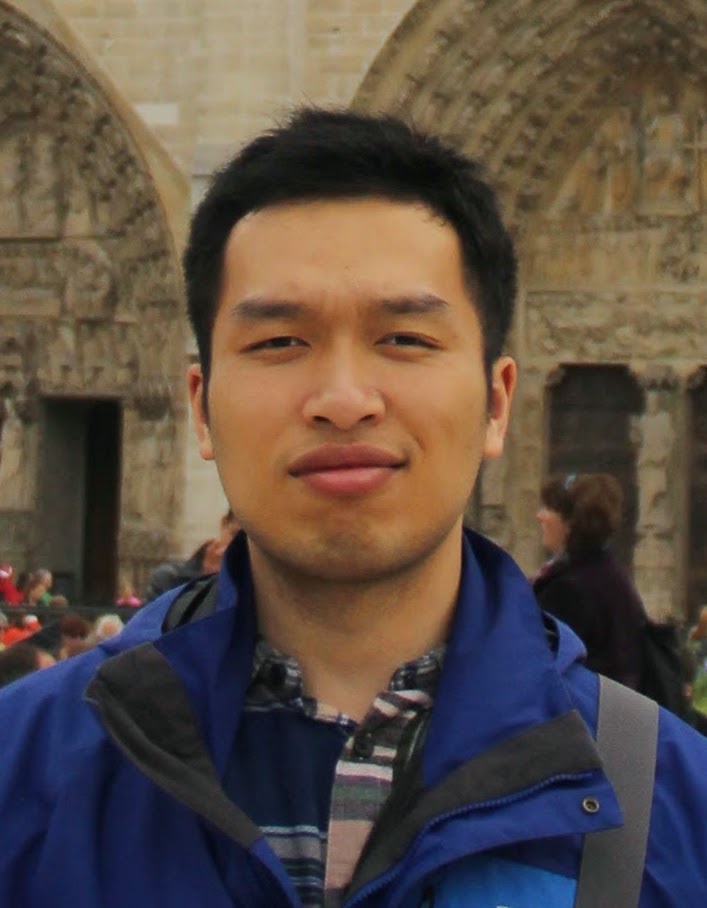}}]{Chenyang Zhu} is an Associate Professor at the School of Computer Science, National University of Defense Technology (NUDT).  He holds both a Bachelor’s and a Master’s degree in Computer Science from NUDT, earned in June 2011 and December 2013, and completed his PhD at Simon Fraser University. His research focuses on computer graphics, 3D vision, robotic perception, and navigation.
\end{IEEEbiography}

\begin{IEEEbiography}[{\includegraphics[width=1in,height=1.25in,clip,keepaspectratio]{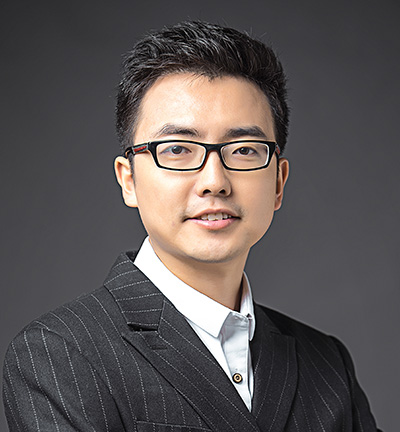}}]{Kai Xu} is a Professor at the School of Computer, National University of Defense Technology, where he received his Ph.D. in 2011. During 2017-2018, he was a visiting research scientist at Princeton University. His research interests include computer graphics, 3D vision and embodied AI. He has published over 100 research papers, including 30+ SIGGRAPH/TOG papers. He serves on the editorial board of ACM Transactions on Graphics, IEEE Transactions on Visualization and Computer Graphics, Computers \& Graphics, Computational Visual Media, and The Visual Computer. He also served as program co-chair and IPC member for several prestigious conferences. His research work can be found in his personal website: www.kevinkaixu.net
\end{IEEEbiography}

\vfill

\end{document}